\pdfoutput=1

\documentclass[journal]{IEEEtran}
\usepackage{tikz,multirow,amsmath,pgfplots,subfigure}
\usetikzlibrary{matrix,chains,positioning,decorations.pathreplacing,arrows}

\usepackage{array}
\usepackage{hyperref}

\newcolumntype{P}[1]{>{\centering\arraybackslash}m{#1}}
\newcolumntype{L}[1]{>{\arraybackslash}m{#1}}

\newcommand{\RNum}[1]{\uppercase\expandafter{\romannumeral #1\relax}}

\ifCLASSINFOpdf
\else
\fi
\begin{document}
\title{EndNet: Sparse AutoEncoder Network for Endmember Extraction and Hyperspectral Unmixing}

\author{Savas~Ozkan,~\IEEEmembership{Member,~IEEE,}
             Berk~Kaya,
	  %Ersin~Esen,~\IEEEmembership{Member,~IEEE,}
            and~Gozde Bozdagi~Akar,~\IEEEmembership{Senior Member,~IEEE}% <-this % stops a space
\thanks{S. Ozkan is with the Image Processing Department, TUBITAK Space Technologies Research Institute and the Department of Electrical and Electronics Engineering, Middle East Technical University, Ankara, Turkey e-mail: savas.ozkan@tubitak.gov.tr}
\thanks{B. Kaya and G.B. Akar are with the Department of Electrical and Electronics Engineering, Middle East Technical University, Ankara, Turkey.}
%\thanks{E. Esen is with the Image Processing Department, TUBITAK Space Technologies Research Institute, Ankara, Turkey}
}

\maketitle

\begin{abstract}
Data acquired from multi-channel sensors is a highly valuable asset to interpret the environment for a variety of remote sensing applications. However, low spatial resolution is a critical limitation for previous sensors and the constituent materials of a scene can be mixed in different fractions due to their spatial interactions. Spectral unmixing is a technique that allows us to obtain the material spectral signatures and their fractions from hyperspectral data. In this paper, we propose a novel endmember extraction and hyperspectral unmixing scheme, so called \textit{EndNet}, that is based on a two-staged autoencoder network. This well-known structure is completely enhanced and restructured by introducing additional layers and a projection metric (i.e., spectral angle distance (SAD) instead of inner product) to achieve an optimum solution. Moreover, we present a novel loss function that is composed of a Kullback-Leibler divergence term with SAD similarity and additional penalty terms to improve the sparsity of the estimates. These modifications enable us to set the common properties of endmembers such as non-linearity and sparsity for autoencoder networks. Lastly, due to the stochastic-gradient based approach, the method is scalable for large-scale data and it can be accelerated on Graphical Processing Units (GPUs). To demonstrate the superiority of our proposed method, we conduct extensive experiments on several well-known datasets. The results confirm that the proposed method considerably improves the performance compared to the state-of-the-art techniques in literature.

\end{abstract}


% \fi
\IEEEpeerreviewmaketitle

\section{Introduction}

\IEEEPARstart{I}{n} remote sensing, hyperspectral data is an essential imaging sensory output by which we gain insight into the Earth system by utilizing information beyond the human visible spectrum. This sensor type has been widely used in a variety of remote sensing applications from environmental monitoring to military surveillance for several decades~\cite{bioucas2012hyperspectral,brown2010hydrothermal}. However, spatial resolution of hyperspectral sensors is very limited compared to other optics operating in visible spectrum domain. Hence, the pixels can be composed of mixture of material spectra. For further process of hyperspectral applications, the constituent materials (endmembers) and their fractions (abundances) should be determined from data.

The mixture of constituent materials is generally formulated with a linear model~\cite{hapke1981bidirectional} wherein each data point on an image  $\mathbf{x} \in \rm I\!R^D$ is a linear combination of endmembers $\mathbf{E}=\left[\mathbf{e}_1, \mathbf{e}_2, ..., \mathbf{e}_K\right] \in \rm I\!R^{D \times K}$ with different fractions $\mathbf{y} =\left[y_1, y_2, ..., y_K\right] \in \rm I\!R^K$. Here, $K$ denotes the number of endmembers and $D$ is the spectral bands of the hyperspectral data. In addition, this formulation has an additional term $\eta$ (it is assumed to possess a zero mean Gaussian noise) to simulate possible noise sources in the process such as sensor readout noise or illumination variability due to surface topography:
\begin{eqnarray}
\label{eqn:bc1}
\mathbf{x} = \sum_{k=1}^{K} {\mathbf{e}_k y_k} + \eta = \mathbf{E}\mathbf{y} + \eta, \hspace{2mm}  s.t. \hspace{1.2mm} y_k \geq 0,\hspace{1.2mm}  \sum_{k=1}^{K} {y_k} = 1
\end{eqnarray}

Unmixing of hyperspectral data is separated into two major steps (note that we assume that the optimum number of endmembers is known) as endmember extraction and quantifying the abundances of these endmembers per pixel. These two unknown variables can be solved either simultaneously or individually depending on the approaches.

In literature, linear and nonlinear models have been vastly studied and several promising methods have been proposed as follows.

For linear models, geometrical volume-based algorithms are quite common to identify endmembers from hyperspectral data~\cite{nascimento2005vertex, boardman1994geometric, harsanyi1994hyperspectral, winter1999n}. They treat the distribution of data samples as a simplex set~\cite{boardman1994geometric}. They ultimately embrace the fact that the vertices of this simplex set correspond to the endmembers since all of the data samples can be spanned with these points. 

By exploiting this assumption (i.e., simplex set) and the notion of the presence of pure pixels, the most straightforward linear solution is to determine unmixed pixels from data for all materials~\cite{winter1999n, heylen2011non}. Vertex Component Analysis (VCA)-like methods~\cite{nascimento2005vertex, harsanyi1994hyperspectral} can extend the assumption by projecting and imposing an orthogonality condition onto endmembers in their estimations. However, these methods have severe drawbacks~\cite{iordache2011sparse}. Briefly, the mixture of materials in both macroscopic and microscopic levels~\cite{iordache2011sparse, singer1979mars} can lead to erroneous spectral estimates. The assumption can collapse with the lack of the purities of one or more materials in data. To solve these  drawbacks, several concepts have been introduced in the literature~\cite{ifarraguerri1999multispectral, wang2010nonnegative, haertel2005spectral, miao2007endmember, li2008minimum} that basically add a margin to the derivation with extra constraints or a kernel structure. This produces more reliable outputs even if the data does not directly satisfy the purest material condition. However, particularly for~\cite{miao2007endmember, li2008minimum}, presence of outliers and noise can significantly reduce the performance of the suboptimal solution of the methods~\cite{jointbayes2009}.

Lastly, methods utilizing a codebook of endmembers (collected with a spectrometry and available in a spectral library)~\cite{iordache2011sparse, iordache2014collaborative, brown2006spectral} are another solution to identify fractional abundances and corresponding endmembers from hyperspectral data. The observation throughout our paper, the importance of sparsity is particularly emphasized in these studies. 

Although the linear mixture model is simple and provides practical advantages, it is not adequate to solve the mixing problem. Multiple scattering effects~\cite{somers2009nonlinear}, microscopic-level material mixtures~\cite{iordache2011sparse, singer1979mars} and water-absorbed environments~\cite{keshava2002spectral} can be shown as severe natural cases that cannot be handled by the linear models. One way to deal with such scenarios is to use unsupervised nonlinear projections of data~\cite{roweis2000nonlinear, bachmann2006improved}. However, this type of approaches requires a high computational workload, thus it eventually reduces the scalability of the methods for large-scale data. An alternative strategy is to replace the conventional operations with a nonlinear kernel function $f(.)$ such that $\mathbf{x} = f(\mathbf{E}\mathbf{y}) + \eta$. This methodology simply introduces a distortion onto endmembers and abundance estimates to enhance the robustness against nonlinear interactions~\cite{fastkernel2016, broadwater2009comparison, chen2013nonlinear, fevotte2015nonlinear, close2012using, altmann2011bilinear, heylen2015nonlinear, kizel2017stepwise, nonlinear2014, super2012}. 

Leveraging supervised data with neural-network architectures is another solution proposed for this problem~\cite{plaza2009use, guilfoyle2001quantitative, plaza2007joint, ayerdi2016, vcanet2016}. Although critical discussions are presented for optimum training sample selection, the quality of training samples drastically influences the performance. Also, data labeling is costly and impractical compared to the unsupervised approaches. The endmembers and fractional abundances should be blind. A recent study~\cite{frosti2017} proposes a blind endmember extraction method based on neural networks by introducing a set of layer components to the encoder layer of an autoencoder. This approach particularly helps to increase the sparsity of fractional abundance estimates. However, no detailed experimental result is presented to validate the superiority of neural networks over the conventional methods and the model is currently suboptimal due to the tendency of overfitting.

The adaptation of neural network structure to hyperspectral unmixing has not been completely solved in previous works, especially for unsupervised setup. To the best of our knowledge, our proposed method will be the first successful attempt based on an autoencoder network that outperforms the conventional methods, all in an unsupervised manner. 

\noindent
\textbf{Our Contribution}: In this study, we propose a novel two-staged neural network autoencoder and an end-to-end learning scheme that are specialized to extract endmembers with their fractional abundances from hyperspectral data. Our method uses some of the major approaches previously introduced for hyperspectral unmixing, however it significantly differs from the traditional autoencoder pipeline with novel features. 

Similar to~\cite{miao2007endmember, li2008minimum}, we observed that rather than strictly parameterized all of the physical properties of the endmembers, solving the problem with some of the constraints can yield better results. Since the spectral angle distance (SAD) is used in our loss function which enforces the maximum angular similarity between samples rather than minimizing the Euclidian distance, the value range of the estimated endmembers is not constrained to [0,1]. However the negative values for endmembers are penalized by a SAD metric in the loss function, thus non-negativity constraint is automatically applied in the learning process. Other constraints such as sum-to-one constraint (i.e., $\sum_{k=1}^{K} {y_k} = 1$) and greater than zero constraint (i.e., $y_k \geq 0$) are preserved because of the internal characteristics of the layer components in the architecture which will be explained in the following section.

In the proposed method, first, sparsity and non-linearity are effectively applied by a rectified activation function, ReLU (i.e., bounds the negative responses of hidden abstracts)~\cite{nair2010rectified} and a normalization layer~\cite{ioffe2015batch} as in~\cite{frosti2017}. This adaptation enables us to set the bias terms to zero in the formulation and it eliminates the adverse effects of bias terms in the solution. However, the complete sparsity is still missing and it is prone to overfitting due to the fact that the network tends to generate hidden abstract (i.e., hidden representation of the neural network model) constantly for highly correlated materials. In other words, a small number of responses should contribute to the composition of a pixel for a practical solution. In our paper, this is achieved by hard-response selection (i.e., i.e., selection of only top responses of the hidden abstracts) and regularization techniques~\cite{srivastava2014dropout, wager2013dropout} which allow only the highest responses of the fractional abundances contribute to the reconstruction stage. To this end, this leads to a better solution and improves the sparsity for fractional abundances. 

 Second, we replace the inner product operators at the encoder layer with spectral angle distance (SAD) to obtain more discriminative hidden abstracts. Third, we introduce an extra set of penalty terms to the loss function in which they can enforce closer angular similarity between the original and the reconstructed data along with the standard Euclidean ($l2$) reconstruction penalty. Moreover, sparse hidden representations can be achieved with additional penalty terms. Fourth, in contrast to some conventional unmixing approaches~\cite{nascimento2005vertex, winter1999n, fevotte2015nonlinear, heylen2015nonlinear, kizel2017stepwise}, our proposed method solves the problem by optimizing endmembers and their corresponding fractional abundances concurrently with a stochastic gradient-based solver. This leads to the optimum solutions for both endmembers and fractions while it scales the method for large-scale data as stated in~\cite{bottou2010large}.

Lastly, we observed that usage of VCA-like~\cite{nascimento2005vertex, winter1999n, heylen2015nonlinear} methods as an autoencoder filter initializer helps us to attain state-of-the-art performance. This makes the filters converge faster to the global optimum. Note that this step is practiced previously in different optimization-based algorithms~\cite{li2008minimum} and it similarly avoids the parameters which leads to poor quality solutions.

The remainder of our paper is organized as follows. First, necessary information about the proposed architecture and the optimization step are presented in Section 2. Section 3 is reserved for the results and discussions on the experiments conducted on several publicly available datasets. Conclusion and overall discussions are presented in Section 4.

\section{Spectral Unmixing}

\noindent
\textbf{Motivation:} A neural network autoencoder is simply composed of two consecutive information processing layers. Encoder layer transforms input samples to hidden abstracts (hidden abstracts correspond to the responses of hidden layers) while the decoder layer tries to restore the original inputs from these hidden abstracts with high accuracy. Ultimately, latent correlations in data are unveiled in this chain of the transformations and conserved as trainable parameter sets in the model~\cite{vincent2010stacked, bengio2013representation, lecun2015deep, rifai2011contractive}. 

For a given sample $\mathbf{x} \in \rm I\!R^D$, an autoencoder initially maps it into a hidden representation $\mathbf{y}\footnote{$\mathbf{y}$ intuitively corresponds to fractional abundances per pixel in hyperspectral unmixing.}  \in \rm I\!R^K$ with the inner product of a trainable parameter set $\mathbf{W}$, $\mathbf{b}$. Following this, the reconstruction of the input sample $\mathbf{\hat{x}} \in \rm I\!R^D$ is recomputed: 
\begin{subequations}
\begin{equation}
\label{eqn:auto1}
\mathbf{y} = f(\mathbf{W}^{(e)}  \mathbf{x} + \mathbf{b}^{(e)}),
\end{equation}
\begin{equation}
\label{eqn:auto2}
{\mathbf{\hat{x}}} = f(\mathbf{W}^{(d)} \mathbf{y}+ \mathbf{b}^{(d)}).
\end{equation}
\end{subequations}

\noindent
Here $f(.)$ indicates an element-wise nonlinear activation function and a logistic activation function is frequently selected as either sigmoid or tanh.

\definecolor{ryb1}{RGB}{141, 211, 199}
\definecolor{ryb2}{RGB}{255, 1, 179}
\definecolor{ryb3}{RGB}{190, 186, 218}
\definecolor{ryb4}{RGB}{251, 128, 114}
\definecolor{ryb5}{RGB}{128, 177, 211}
\definecolor{ryb6}{RGB}{253, 180, 98}
\definecolor{ryb7}{RGB}{179, 222, 105}
\definecolor{ryb8}{RGB}{63, 122, 205}
\definecolor{ryb9}{RGB}{243, 65, 205}

\begin{figure}[t!]
\centering
\subfigure{
\begin{tikzpicture}
  \begin{axis} [
      width=4.9cm, height=3.2cm,
      xlabel = \footnotesize $Spectral Bands$,
      ylabel = {\footnotesize $Reflectance$},
      ylabel style={yshift=-0.55cm},
      xlabel style={yshift=0.2cm},
      xticklabel style = {font=\footnotesize},
      yticklabel style = {font=\footnotesize}
]
    \addplot [thick,ryb1] table {indian_data_e1.txt};
    \addplot [thick,ryb2] table {indian_data_e2.txt};
    \addplot [thick,ryb3] table {indian_data_e3.txt};
    \addplot [thick,ryb4] table {indian_data_e4.txt};
    \addplot [thick,ryb5] table {indian_data_e5.txt};
    \addplot [thick,ryb6] table {indian_data_e6.txt};
    \addplot [thick,ryb7] table {indian_data_e7.txt};
    \addplot [thick,ryb8] table {indian_data_e8.txt};

;
  \end{axis}
\end{tikzpicture}}
\subfigure{
\begin{tikzpicture}
  \begin{axis}[
      width=4.9cm, height=3.2cm,
      xlabel = \small $Spectral Bands$,
      xlabel style={yshift=0.2cm},
      yticklabel style = {font=\footnotesize},
      xticklabel style = {font=\footnotesize},
]
    \addplot [thick,ryb9] table {indian_data_b.txt};
  \end{axis}
\end{tikzpicture}}

\caption{Estimates of a conventional autoencoder for hyperspectral unmixing problem. First plot (left side) shows decoder layer parameters, $\mathbf{W}^{(e)}$, which intuitively correspond to the endmembers while second plot (right side) illustrates bias term response. As can be seen from the results, bias term acts like one of the endmembers exhibited from the scene. Moreover, the estimated signatures (left side) do not satisfy the physical conditions of endmembers, i.e., $y_k \geq 0$. }
\label{fig:conauto}
\end{figure}
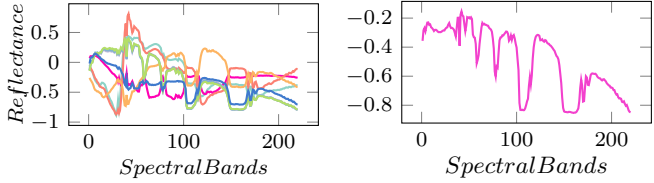

Furthermore, some structural variations are observed for activation functions and/or parameter sets in literature~\cite{bengio2013representation, lecun2015deep, rifai2011contractive, rasmus2015semi}. More precisely, the activation function $f(.)$ at the decoder layer can be discarded or same/disjoint parameter sets can be used for better error propagation.

Finally, the parameters are optimized by minimizing the standard Euclidean ($l2$) reconstruction error of the original and the reconstructed samples:
\begin{eqnarray}
\label{eqn:bc2}
\mathcal{L} = \frac{1}{2}\|\mathbf{x} - \mathbf{\hat{x}} \|_2^2.
\end{eqnarray}

\noindent
where $\mathcal{L}$ is the loss function. Fig.~\ref{fig:conauto} presents the possible parameter outcomes of a conventional autoencoder learned by Eqs.(\ref{eqn:auto1}), (\ref{eqn:auto2}) and (\ref{eqn:bc2}) on the University of Pavia dataset~\cite{holzwarth2003hysens}.

\noindent
\textbf{Current Limitations For Unmixing Problem:} The direct use of a traditional autoencoder might not yield a stable solution particularly for several reasons: 1) Logistic functions tend to generate redundant responses even if the input samples are not correlated with the parameter set~\cite{nair2010rectified, he2015delving, frosti2017}. This also violates and decreases the sparsity of true projections. 2) The inner product is not sufficiently discriminative and it can lead to erroneous fractional abundances for data. 3) Although the standard Euclidean ($l2$) reconstruction term is the main penalty function used to minimize the estimation error, it should be carefully regularized with extra terms related to the domain in order to reach to the optimum solution. Otherwise, the solution can be stuck to a local minimum. 

For these reasons, these current limitations aggravate conventional autoencoders to attain state-of-the-art performance on hyperspectral unmixing problem.

\subsection{Sparse Autoencoder for Hyperspectral Unmixing}

The primary objective of this paper is to accomplish the adaptation of an autoencoder pipeline to the endmember extraction for hyperspectral data. In particular, we strive to preserve the estimation power of autoencoders on latent data correlations as well as the observations that are previously introduced for endmember extraction. 

To enhance the performance and make the autoencoder effective for the problem, we made several modifications in the architecture and the optimization step\footnote{Source code and presented results will be available on \url{https://github.com/savasozkan/endnet} } as follows. First, bias terms are removed from the formulations due to their adverse effects in the problem. Second, we replace and introduce additional layers to the architecture to improve sparsity as well as consistency. Later, inner product is replaced with spectral angle distance (SAD) to estimate more discriminative fractional abundances. Finally, we propose a novel loss function to optimize the model parameters effectively. In the following subsections, we describe these modifications in details.

\noindent
\textbf{Zero-biased Filters}: In neural network architecture, bias term is one of the critical parameters, especially for linear regression models~\cite{memisevic2014zero}, since it determines the influence of parameters for next layers by thresholding the responses through activation functions. However, bias terms violate one of the crucial properties of endmembers by which none of the materials can be constantly expected in every pixel compositions, if a scene is not composed of one material.

More clearly, since bias terms constantly affect the composition of pixels, they behave like one of the endmembers observed from the data at the end of the optimization step. Fig.~\ref{fig:conauto}. plots a possible bias term response for a conventional autoencoder. For this reason, we removed the bias terms in both Eqs.(\ref{eqn:auto1}) and (\ref{eqn:auto2}) for our model. Also, this step partially retains the simplex set assumption (i.e., an affine projection) which constitutes a basis for the conventional methods in the literature~\cite{nascimento2005vertex, boardman1994geometric, harsanyi1994hyperspectral}.

Moreover,  we discarded the activation function at the decoder layer to propagate the error more effectively and intuitively $\mathbf{W}^{(d)}$ parameter begins to represent endmembers $\mathbf{E}$ estimated from the scene. Similarly, hidden abstract $\mathbf{y}$ corresponds to the fractional abundances of a pixel. 

Lastly, since abundance estimates possess non-linear relations and cannot be computed directly from endmembers $\mathbf{W}^{(d)}$, we used disjoint parameter sets for both encoder and decoder layers  ($\mathbf{W}^{(e)}  \in \rm I\!R^{K \times D}$ and $\mathbf{W}^{(d)} \in \rm I\!R^{D \times K}$). 

\noindent
\textbf{Sparsity and Nonlinearity}:  Sparsity and nonlinearity are critical features for endmember extraction and hyperspectral unmixing~\cite{iordache2011sparse, iordache2014collaborative, chen2013nonlinear, fevotte2015nonlinear} in order to make a robust estimation. As previously mentioned limitations, logistic functions are insufficient to conserve the sparsity for the architecture even if they generate nonlinear responses. Therefore, we made three critical modifications in the architecture to achieve sparse, consistent and nonlinear hidden abstracts.

\begin{figure}[t]
\centering

\includegraphics[scale=0.22]{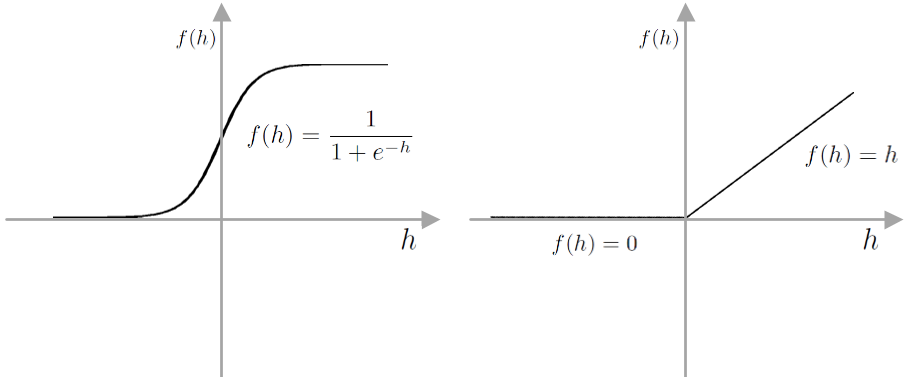}
  \caption{Response characteristics of Sigmoid (left) and ReLU (right) activation functions.}

\label{fig:relu}
\end{figure}

First,  a rectified linear activation~\cite{nair2010rectified} is used at the encoder layer instead of logistic functions. Fig.~\ref{fig:relu} illustrates the response characteristics of the Sigmoid and Rectified Linear Unit (ReLU) activations for different input values.  ReLU bounds the negative influence of filters and increases the selectivity of the representation. In addition, it boosts the parameters and allows them to saturate more quickly in the optimization step. In the scope of this paper, we tested our proposed method with the variants of rectified activation units~\cite{he2015delving, clevert2015fast, xu2015empirical} to be sure about the optimum solution. These activation units permit limited amount of negative responses to the subsequent layers. However, these negative values ultimately perturb the relation of endmembers (i.e. they become more correlated) and violates the greater than zero constraint (i.e., $y_k > 0$) for the abundance estimates.

\newcommand\gauss[2]{1/(#2*sqrt(2*pi))*exp(-((x-#1)^2)/(2*#2^2))}

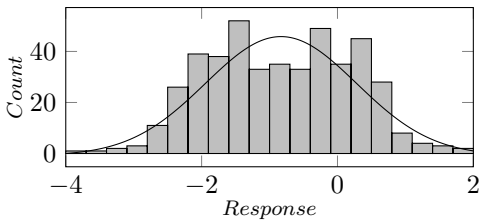
\begin{figure}[b!]
\centering

\subfigure{
\begin{tikzpicture}
    \begin{axis}[
        height=3.7cm,
        width=7cm,
      xlabel = \footnotesize $Response$,
      ylabel = {\footnotesize $Count$},
      ylabel style={yshift=-0.55cm},
      xlabel style={yshift=0.2cm},
        xmin=-4,
        xmax=2,
    ]

    \addplot[
        black,
        fill=lightgray,
        hist,
        hist/bins=20,
    ] table[
        y=true-error,
    ] {data_bn.txt};

\addplot[domain={-4:2},yscale=130,samples=448] {\gauss{-0.8315}{1.1127}};
    \end{axis}
\end{tikzpicture}}
\caption{Distribution characteristic of batch normalization layer outputs. Due to shifting parameter $\boldsymbol{\rho}$, mean value of the distribution slightly moves to left in the plot.}

\label{fig:batch}
\end{figure}

However, trainable parameters can quickly become ill-posed (i.e. small changes in the parameters can lead to oscillation and instability in the model) with the removal of the logistic function~\cite{ioffe2015batch, krizhevsky2012imagenet}. Therefore, we introduced a normalization layer~\cite{ioffe2015batch} before the ReLU activation function at the encoder layer. For our model, this layer especially induces selectivity on the top activation responses (it has a similar objective as a bias term~\cite{memisevic2014zero}) in addition to the mitigation of ill-posed effects. It normalizes the responses by scaling and reordering them. Fig.~\ref{fig:batch} illustrates the possible output distribution characteristics of this normalization layer. If we assume that shifting parameter $\boldsymbol{\rho}$ in Eq.(\ref{eqn:bn}) is discarded, by combining with ReLU layer, it would permit roughly 50\% of the top positive responses to the next layer at each iteration: 
\begin{eqnarray}
\label{eqn:bn}
\text{BN}(\mathbf{h})  = \frac{ (\mathbf{h} - \boldsymbol{\mu}) }{\sqrt{\boldsymbol{\sigma}^2 + \epsilon } } + \boldsymbol{\rho},
\end{eqnarray}

\noindent
where  $\text{BN}(.)$ is the output of the normalization layer and it is fed into ReLU layer in Eq.(\ref{eqn:act}). $\epsilon$ is a very small constant ($10^{-8}$) to prevent zero-division in the formulation. $\mathbf{h} = \mathbf{W}^{(e)} \mathbf{x}$, $\mathbf{h} \in \rm I\!R^K$ is the filter response for input $\mathbf{x}$  (We should note that we will redefine this filter response in the following subsection. This formulation is given here to preserve the consistency). $\boldsymbol{\mu}= \frac{1}{N}\sum_{i=1}^{N} {\mathbf{x}^{(i)}}, \boldsymbol{\mu} \in \rm I\!R^{K}$ and $\boldsymbol{\sigma}^2 = \frac{1}{N} \sum_{i=1}^{N} {(\mathbf{x}^{(i)} - \boldsymbol{\mu})^2}, \boldsymbol{\sigma}^2 \in \rm I\!R^{K}$ are the mean and variance of mini-batch filter responses at $t^{th}$ iteration. $\boldsymbol{\rho} \in \rm I\!R^{K}$ are the trainable parameters to adjust the permutability of filter responses. We should remark that scaling factor in~\cite{ioffe2015batch} is discarded due to better fractional abundances and endmember extraction: 
\begin{eqnarray}
\label{eqn:act}
&&f(\mathbf{h}) = \bigg\{ 
\begin{matrix}
\text{BN}(\mathbf{h}), & \text{\parbox{4cm}{if $\text{BN}(\mathbf{h})$ $>$ 0}} \\ 
0, & \text{\parbox{4cm}{otherwise}} \\ 
\end{matrix} 
\end{eqnarray}

For ReLU activation and normalization layers, similar observation (i.e., sparsity and ill-posed artifacts) can be found in the very recent work~\cite{frosti2017}. 

However, autoencoders might not still have the optimum solution in some cases due to parameter overfitting (i.e. lack of full parameter convergence and sparsity). In order to make improvements for reliable abundance estimates and endmembers, we utilized three additional steps at the end of the activation function. For this purpose, we first employed a regularization layer, i.e., Dropout~\cite{srivastava2014dropout, wager2013dropout} with $\mathbf{r}\sim \text{Bernoulli}(p)$, after the activation function $f(.)$ as $\mathbf{z}=\mathbf{r} \ast f(\mathbf{h})$. Ultimately, for each iteration, the network introduces a new solution for the problem by randomly ignoring some of the activations so that the generalization capacity of the network is enhanced. Here $\ast $ denotes an element-wise product and $0<p \le 1$ should be defined by the user based on the spectral correlations of materials in a scene (i.e if highly correlated materials exist in the scene, $p$ should be close to $0.5$). Note that the default value of $p$ is set to $1.0$.

Second, we hardly selected the top $n$ activations (i.e., highest ones)  from $\mathbf{z}$ as in~\cite{makhzani2013k} ($n$ is fixed to 2 due to the optimum spatial material mixture) to increase the selectivity of $\mathbf{W}^{(d)}$. Finally, we applied $l1$ normalization to satisfy the sum-to-one constraint (i.e., $\sum_{k=1}^{K} {y_k} = 1$) on abundance estimates:
\begin{eqnarray}
\label{eqn:l1}
&&\mathbf{y} = \frac{\mathbf{z}^*}{  (\|  \mathbf{z}^*  \|_1   + \epsilon)},
\end{eqnarray}

\noindent
where $\mathbf{z}^*$ denotes $n$-top activations. To learn the parameters in our model, we use backpropagation algorithm~\cite{lecun1989backpropagation} and Eq.(\ref{eqn:l1}) is differentiable except at zero (For this case, partial derivative is directly set to zero). From chain rule, the partial derivative of the $l1$ normalization layer is computed as follows:
\begin{eqnarray}
\label{eqn:bc4}
 \frac{\partial \mathcal{L}}{\partial \mathbf{z}^*} =  \frac{1}{\|  \mathbf{z}^* \|_1}\bigg( \frac{\partial \mathcal{L}}{\partial \mathbf{y}}  - \mathbf{y} \sum_{k=1}^{K} \frac{\partial \mathcal{L}}{\partial y_k} sign(z^*_k) \bigg) .
\end{eqnarray}

\noindent
where $\mathcal{L}$ is the loss function. The term $\frac{\partial \mathcal{L}}{\partial \mathbf{y}}$ defines the propagated error up to the hidden abstracts $\mathbf{y}$ from the upper layer. Lastly, $sign(.)$ denotes a function which returns the sign of its input.

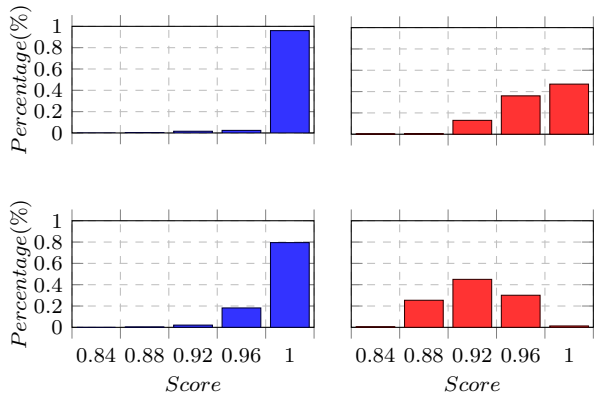
\begin{figure}[b]
\centering

\subfigure{
\label{fig:a}
\begin{tikzpicture} 
\begin{axis}[
     width=4.8cm, height=3.0cm,
      ylabel = {\footnotesize $Percentage(\%)$},
      ylabel style={yshift=-0.55cm},
     xticklabel style = {font=\footnotesize},
     yticklabel style = {font=\footnotesize},
     ybar interval, ymax={1.0},ymin=0.0,  
     xmax={1.04},xmin=0.84,  
     grid style=dashed,
    ymajorgrids=true, ybar interval=0.8, xticklabels={,,}
]
\addplot [blue!20!black,fill=blue!80!white] coordinates  { (0.84, 0.0) (0.88, 0.003) (0.92, 0.016) (0.96, 0.024) (1.0, 0.96) (1.04, 0.96)};
\end{axis}
\end{tikzpicture}}
\subfigure{
\label{fig:b}
\begin{tikzpicture} 
\begin{axis}[
     width=4.8cm, height=3.0cm,
     xticklabel style = {font=\footnotesize},
     yticklabel style = {font=\footnotesize},
     ybar interval, ymax={1.0},ymin=0.0,  
     xmax={1.04},xmin=0.84,  
     grid style=dashed,
    ymajorgrids=true, ybar interval=0.8,
    yticklabels={,,}, xticklabels={,,}
]
\addplot [red!20!black,fill=red!80!white] coordinates  { (0.84, 0.003) (0.88, 0.006) (0.92, 0.13) (0.96, 0.36) (1.0, 0.47) (1.04, 0.475)};
\end{axis}
\end{tikzpicture}}

\subfigure{
\label{fig:b}
\begin{tikzpicture} 
\begin{axis}[
     width=4.8cm, height=3.0cm,
      xlabel = \footnotesize $Score$,
      ylabel = {\footnotesize $Percentage(\%)$},
      ylabel style={yshift=-0.55cm},
     xticklabel style = {font=\footnotesize},
     yticklabel style = {font=\footnotesize},
     ybar interval, ymax={1.0},ymin=0.0, 
     xmax={1.04},xmin=0.84,  
     grid style=dashed,
    ymajorgrids=true, ybar interval=0.8,
     xtick={0.84, 0.88, 0.92, 0.96, 1.0, 1.04}
]
\addplot [blue!20!black,fill=blue!80!white] coordinates  { (0.84, 0.000) (0.88, 0.003) (0.92, 0.021) (0.96, 0.182) (1.0, 0.795) (1.04, 0.795)};
\end{axis}
\end{tikzpicture}}
\subfigure{
\label{fig:b}
\begin{tikzpicture} 
\begin{axis}[
     width=4.8cm, height=3.0cm,
     xlabel = \footnotesize $Score$,
     xticklabel style = {font=\footnotesize},
     yticklabel style = {font=\footnotesize},
     ybar interval, ymax={1.0},ymin=0.0, 
     xmax={1.04},xmin=0.84,  
     grid style=dashed,
    ymajorgrids=true, ybar interval=0.8,
    yticklabels={,,},
     xtick={0.84, 0.88, 0.92, 0.96, 1.0, 1.04}
]
\addplot [red!20!black,fill=red!80!white] coordinates  { (0.84, 0.005) (0.88, 0.254) (0.92, 0.45) (0.96, 0.301) (1.0, 0.013) (1.04, 0.013)};
\end{axis}
\end{tikzpicture}}

\caption{Normalized similarity score distributions for same (blue) and very coherent (red) class comparisons with inner product (first row) and spectral angle distance (second row). As can be seen from the plots, spectral angle distance obtain more discriminative results.}

\label{fig:dist}
\end{figure}

\noindent
\textbf{Discriminative Hidden Abstracts}: For neural network autoencoders, input data samples need to be mapped to hidden abstracts truly by storing their latent correlations, in order to recompute the original ones with high accuracy. This is why, an accurate mapping is of a critical requirement to reveal these latent correlations from data.

Except in minor instances~\cite{chunjie2017cosine}, the inner products of input samples with filter parameters have produced the best performance for almost all applications in the visible domain~\cite{bengio2013representation,  he2015delving, krizhevsky2012imagenet}. For the hyperspectral domain, there are various useful operators which here prove their success on several tasks such as classification and spectral signature comparison~\cite{dennison2004comparison, camps2016kernel, camps2006composite, fauvel2006evaluation, Lihybrid2011}. 

Fig.~\ref{fig:dist} plots the similarity score distributions of the normalized inner product (first row) and the spectral angle distance (SAD) (second row) of same (blue) and very coherent (red) material spectral signatures. From these plots, it is observed that inner product constantly produces high scores even if different materials are compared. However, the same material scores for spectral angle distance (SAD) can be separated from the coherent ones. In addition,~\cite{Lihybrid2011} explains that SAD is better for separability, while spectral information divergence (SID) is more practical for preserving spectral patterns. Note that separability and discriminative power are especially important to estimate sparse responses at the encoder layer.

For these reasons, we replaced inner product with SAD at the encoder layer to obtain more discriminative and separable hidden abstracts as well as endmembers from hyperspectral data.

Spectral angle distance (SAD) measures the spectral angle between two input samples and the score closer to zero implies higher correlation. One of the concerns addressed for the SAD metric~\cite{camps2016kernel, liu2013kernel} is that it is limitation to the nonlinear cases. As a remedy, several improvements are presented in the literature and most of which are based on kernel-based methods~\cite{camps2016kernel, camps2006composite, fauvel2006evaluation}. However, this limitation is not a drawback for our case, since nonlinearity is supplied with the nonlinear activation function~\cite{nair2010rectified} and the normalization layer~\cite{ioffe2015batch} at the output of this metric. 

SAD computes the similarity score of two samples, $\mathbf{x}^{(i)}$ and $\mathbf{x}^{(j)}$, as follows:
\begin{eqnarray}
\label{eqn:sad}
\text{S}(\mathbf{x}^{(i)}, \mathbf{x}^{(j)}) =\text{cos}^{-1} \big( \theta(\mathbf{x}^{(i)}, \mathbf{x}^{(j)}) \big) ,
\end{eqnarray}

\noindent
where
\begin{eqnarray}
\label{eqn:bc4}
\theta(\mathbf{x}^{(i)}, \mathbf{x}^{(j)}) = \frac{\mathbf{x}^{(i)} \mathbf{x}^{(j)}}{ \| \mathbf{x}^{(i)} \|_2 \| \mathbf{x}^{(j)} \|_2}.
\end{eqnarray}

In Eq.(\ref{eqn:sad}), the SAD scores vary between $[0, \pi]$ and zero value means that two samples are identical. Therefore, we first need to normalize the score $\text{S}(.,.)$ to [0, 1], since a higher score means a higher similarity at the decoder layer:
\begin{eqnarray}
\label{eqn:c}
\text{C}(\mathbf{x}^{(i)}, \mathbf{x}^{(j)}) = 1.0 - \frac{\text{S}(\mathbf{x}^{(i)}, \mathbf{x}^{(j)})}{\pi}.
\end{eqnarray}

From this moment on, we reformulate the filter responses as $\mathbf{h} = \text{C}(\mathbf{x},\mathbf{W}^{(e)})$. SAD is differentiable to optimize necessary parameters with the backpropagation algorithm by minimizing the loss function $\mathcal{L}$. By applying chain rule again, the gradient of $k^{th}$ row $\mathbf{w}_k$ of  $\mathbf{W}^{(e)}, \mathbf{w}_k \in \rm I\!R^{1 \times D}$ can be computed as:
\begin{eqnarray}
\label{eqn:der}
 \frac{\partial \mathcal{L}}{\partial \mathbf{w}_k} =  \frac{\partial \mathcal{L}}{\partial \text{C}_k}  \frac{\partial \text{C}_k}{\partial \text{S}_k} \frac{\partial \text{S}_k}{\partial \theta_k} \frac{\partial \theta_k}{\partial \mathbf{w}_k}, \hspace{2mm} k=1,2, ..., K.
\end{eqnarray}

\noindent
where $\text{C}_k$, $\text{S}_k$ and $\theta_k$ are the simplifications of $\text{C}(\mathbf{x},\mathbf{w}_k)$, $\text{S}(\mathbf{x},\mathbf{w}_k)$ and $\theta(\mathbf{x},\mathbf{w}_k)$ respectively to ease the formulation and understandability. Also, each derivation term in Eq.(\ref{eqn:der}) can be written as: 
\begin{subequations}
\begin{equation}
\frac{\partial \text{C}_k}{\partial \text{S}_k} = \frac{-1}{\pi}, 
\end{equation}
\begin{equation}
\frac{\partial \text{S}_k}{\partial \theta_k} =  \frac{-1}{\sqrt{1-\theta^2_k}},
\end{equation}
\begin{equation}
\frac{\partial \theta_k}{\partial \mathbf{w}_k} =  \frac{\mathbf{x}}{\| \mathbf{w}_k \|_2 \| \mathbf{x} \|_2} - \frac{\mathbf{w}_k \ast (\mathbf{w}_k \mathbf{x}) }{\| \mathbf{w}_k \|^3_2 \| \mathbf{x} \|_2}.
\end{equation}
\end{subequations}

In the next subsection, we will analyze the optimization step of the proposed method by introducing modifications in the objective function. 

\noindent
\textbf{Learning}: Parameter optimization aims to minimize the error between input samples and their reconstructed versions by learning useful latent correlations from data. However, we found out that use of the Euclidean ($l2$) norm as a primary objective term to unveil these correlations is not completely adequate for the problem. The main limitation of Euclidean norm is that it aggravates the method by estimating  inappropriate/underestimated endmembers under severe illumination changes and nonlinearity (i.e., spectral variability)~\cite{tsai1998derivative}. On the other hand, the spectral angle distance-like (SAD) operator overcomes these limitations and improves unmixing performance by exploiting geometric features of samples as explained in~\cite{kizel2017stepwise, rabah2011new}. In addition, smoothness and sparsity priors should be considered in the loss function for better parameter convergence. For this purpose, we made a critical set of modifications in the objective function related to this domain and a novel objective function $\mathcal{L}$ is reformulated as:
\begin{eqnarray}
\label{eqn:bc5}
\begin{aligned}
\mathcal{L} = \frac{\lambda_0}{2}\|\mathbf{x} -\mathbf{\hat{x}} \|_2^2 - \lambda_1 \text{D}_{\text{KL}}\big( \hspace{1mm} 1.0 || \text{C}(\mathbf{x}, \mathbf{\hat{x}}) \hspace{1mm}  \big) + \lambda_2 \|  \mathbf{z} \|_1 \\ 
+ \lambda_3 \| \mathbf{W}^{(e)} \|_2 + \lambda_4 \| \mathbf{W}^{(d)} \|_2 +  \lambda_5 \| \boldsymbol{\rho} \|_2
\end{aligned}
\end{eqnarray}

\noindent
where $C(.,.)$ is the normalized SAD score between the original and reconstructed version as in Eq.(\ref{eqn:c}). $\lambda_0$ is the term that controls the influence of Euclidean norm. Additionally, we add a Kullback-Leibler divergence term ($D_{KL}$)~\cite{hershey2007approximating, maaten2008visualizing} to maximize SAD score distributions between the original and the reconstructed samples. $\lambda_1$ determines the effect of this cost term to the objective function. Note that Euclidean norm term is still critical for stable estimates and it cannot be directly set to zero, contrary to~\cite{frosti2017}. Thus, we empirically set $\lambda_0$ and $\lambda_1$ to $0.01$ and $10$ respectively.

As stated, the sparsity is important for effective hyperspectral unmixing. Therefore, we introduce a $l1$ regularization term $\|  \mathbf{z} \|_1$ as in~\cite{kavukcuoglu2010fast} which penalizes the hidden layers that constantly generate responses for different input samples. Thus, the sparsity is enforced to the hidden abstracts and more distinct filter parameters are obtained for each material. The influence of $l1$ sparsity term $\lambda_2$ is tuned to $0.1$ which is a relatively large value. But, we should note that this value needs to be increased/decreased depending on the mixture level of the scenes.

We also add $l2$ smoothing terms to the objective function for each trainable parameter and we define their values as $10^{-5}$, $10^{-5}$ and $10^{-3}$ for $\lambda_3$, $\lambda_4$ and $\lambda_5$ respectively. 

The model parameters are optimized with a gradient-based stochastic Adam~\cite{kingma2014adam} optimizer by minimizing the loss function. We fixed the learning rate and number of training iterations to $0.001$ and $400K$ respectively for all datasets (These terms can be still tuned). Additionally, mini-batch size $N$ is set to 64 to balance the accuracy and the complexity at each iteration. Unlike the suggested values for momentum terms as in~\cite{kingma2014adam}, we empirically found out that $\beta_1$ value should be defined as $0.7$ to reduce the oscillation and instability of learning in the model.

\begin{figure}[t]
\centering

\subfigure{
\label{fig:a}
\includegraphics[scale=0.36]{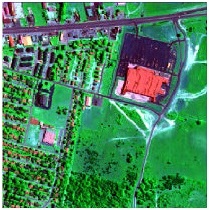}}
\subfigure{
\label{fig:b}
\includegraphics[scale=0.36]{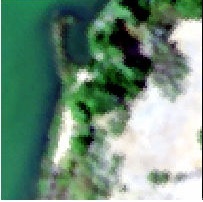}}
\subfigure{
\label{fig:c}
\includegraphics[scale=0.36]{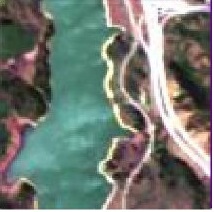}}
\subfigure{
\label{fig:d}
\includegraphics[scale=0.36]{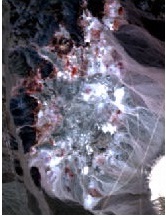}}

\caption{Four real hyperspectral datasets used in the experiments: Urban, Samson,  Jasper Ridge and Cuprite respectively. These datasets are broadly used to evaluate the performance of a method for endmember extraction and abundances unmixing o hyperspectral data.}

\label{fig:data1}
\end{figure}

Lastly, the parameters ($\mathbf{W}^{(e)}$ and $\mathbf{W}^{(d)}$) are initialized with the estimates of VCA-like methods~\cite{nascimento2005vertex, winter1999n, heylen2015nonlinear} instead of a random initialization. This approach provides several advantages: First, it is practical to start the optimization from a more reliable initialization which can span all data points to decrease the possibility of parameter overshooting  (if any geometrical volume-based algorithm is used). Second, it speeds up the convergence of the parameters. 

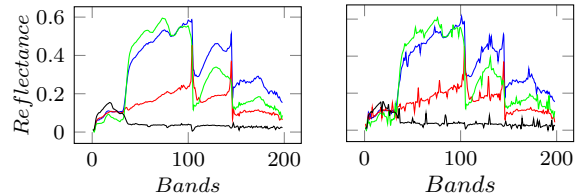
\begin{figure}[b!]
\centering

\subfigure{
\begin{tikzpicture}
  \begin{axis} [
      width=4.6cm, height=3.4cm,
      xlabel = \footnotesize $Bands$,
      ylabel = {\footnotesize $Reflectance$},
      ylabel style={yshift=-0.55cm},
      xlabel style={yshift=0.2cm},
      xticklabel style = {font=\footnotesize},
      yticklabel style = {font=\footnotesize}
]
    \addplot [smooth,blue] table {data_o1.txt};
    \addplot [smooth,red] table {data_o2.txt};
    \addplot [smooth,green] table {data_o3.txt};
    \addplot [smooth,black] table {data_o4.txt};
  \end{axis}
\end{tikzpicture}}
\subfigure{
\begin{tikzpicture}
  \begin{axis}[
      width=4.6cm, height=3.4cm,yticklabels={,,},
      xlabel = \small $Bands$,
      xlabel style={yshift=0.2cm},
      xticklabel style = {font=\footnotesize},
]
    \addplot [smooth,blue] table {data_d1.txt};
    \addplot [smooth,red] table {data_d2.txt};
    \addplot [smooth,green] table {data_d3.txt};
    \addplot [smooth,black] table {data_d4.txt};
  \end{axis}
\end{tikzpicture}}

\caption{Four samples corrupted with the denoising criterion. Different color is used for each hyperspectral signature pair.}

\label{fig:corr}
\end{figure}

\noindent
\textbf{Prevent Autoencoder From Overfitting:} As mentioned, we utilized Dropout~\cite{srivastava2014dropout, wager2013dropout} at the hidden layer to prevent hidden abstracts from overfitting to a poor quality solution.  In addition, we employ denoising autoencoder scheme~\cite{vincent2010stacked} in the parameter optimization step to be robust to noise exhibited on data (i.e. approximation error $\eta$ in the process). Even though the primary principles for both methods are similar, we used these methods at different layers to generalize our model parameters.

This scheme is essentially based on the fact that data samples are initially corrupted with an additive Gaussian noise on purpose while the reconstruction error is computed over the original ones. This helps to improve the generalization capacity of the encoder parameter set and it leads to better solution for robust representations by considering possible deformations (i.e. illumination changes, sensor noise etc.) in advance. 

\begin{table*}[t]
\label{tab:load}
\begin{center}

\caption{SAD and RMSE results on Urban dataset. Mean and standard deviation are reported. Best results are illustrated in bold.}
\label{table:urban}

\begin{tabular}{P{0.8cm} || P{1.45cm} P{1.45cm} P{1.45cm} P{1.45cm} P{1.45cm}  P{1.45cm}  P{1.45cm} P{1.45cm} P{1.45cm} }
  \hline
   \multirow{2}{*}{Endm.} &  \multicolumn{9}{c}{Spectral Angle Distance (SAD) ($\times 10^{-2}$)} \\ \cline{2-10}
      &  VCA  & DMaxD &  MVSA & SPICE & SCM & $l_{1|2}$-NMF & DgS-NMF & EndNet-VCA  & EndNet-DMaxD \\ \hline
     \#1   &  21.56 $\pm$3.1 &  13.16 $\pm$0.0 &  19.04 $\pm$0.1 &  45.04 $\pm$ $>$9 &  9.38 $\pm$0.0 &  6.06 $\pm$0.2 & \textbf{5.86 $\pm$ 0.1} & 6.72 $\pm$0.2 & 6.88 $\pm$0.2 \\
     \#2   &  39.04 $\pm$5.1  &  $>$99 $\pm$0.0 &  $>$99 $\pm$1.3 &  $>$99 $\pm$$>$9 &  10.55 $\pm$0.0 & 20.05 $\pm$0.4 & 13.69 $\pm$0.4 & 4.04 $\pm$0.1  & \textbf{3.92 $\pm$0.3} \\
     \#3   &  26.77 $\pm$7.5  &  7.43 $\pm$0.0 &  37.83 $\pm$9.4 &  30.89 $\pm$$>$9 &  14.03 $\pm$0.0 & 3.71 $\pm$0.0 &  4.12 $\pm$0.0  & 3.68 $\pm$0.2  & \textbf{3.53 $\pm$0.1} \\
     \#4   &  82.39 $\pm$0.1  &  21.74 $\pm$0.0 &  18.41 $\pm$1.9 &  66.37 $\pm$$>$9 &  41.86 $\pm$0.0 & 14.22 $\pm$0.2 & 10.54 $\pm$0.3  & 3.99 $\pm$0.7 & \textbf{3.35 $\pm$0.5} \\ \hline
     Avg.  &  41.77 $\pm$4.5  &  37.88 $\pm$0.0 &  43.81 $\pm$3.1 &  60.57 $\pm$$>$9 &  18.79 $\pm$0.0 & 11.01 $\pm$0.2 & 8.55 $\pm$0.2  & 4.60 $\pm$0.3  & \textbf{4.42 $\pm$0.3} \\
  \hline \hline

   \multirow{2}{*}{} &  \multicolumn{9}{c}{Root Mean Square Error (RMSE) ($\times 10^{-2}$)} \\ \cline{2-10}
      &  VCA  & DMaxD &  MVSA & SPICE & SCM & $l_{1|2}$-NMF & DgS-NMF & EndNet-VCA  & EndNet-DMaxD \\ \hline
     \#1   &  31.31 $\pm$7.2  &  27.08 $\pm$0.0 &  27.69 $\pm$0.3 &  29.43 $\pm$3.4 &  32.79 $\pm$0.0 &  14.77 $\pm$0.1 & 13.18 $\pm$0.1 & \textbf{10.24 $\pm$0.1} & 10.41 $\pm$0.2 \\
     \#2   &  41.98 $\pm$5.6  &  48.99 $\pm$0.0 &  31.07 $\pm$0.0 &  35.32 $\pm$4.5 &  36.25 $\pm$0.0 & 16.16 $\pm$0.2 & 12.95 $\pm$0.0 & 13.01 $\pm$0.3  & \textbf{12.24 $\pm$0.3} \\
     \#3   &  27.93 $\pm$3.7 &  28.07 $\pm$0.0 &  27.27 $\pm$0.4 &  26.33 $\pm$1.9 &  32.61 $\pm$0.0 & 12.65 $\pm$0.2 &  9.57 $\pm$0.1  & 8.69$\pm$0.3  & \textbf{8.35 $\pm$0.3} \\
     \#4   &  21.91 $\pm$2.1 &  21.59 $\pm$0.0 &  21.13 $\pm$0.4 &  21.93 $\pm$6.9 &  32.86 $\pm$0.0 & 6.90 $\pm$0.1 & 6.27 $\pm$0.0  & 6.13 $\pm$0.2 & \textbf{5.92 $\pm$0.1} \\ \hline
     Avg.  &  30.79 $\pm$4.7  &  31.43 $\pm$0.0 &  26.79 $\pm$0.4 &  28.25 $\pm$4.2 &  33.59 $\pm$0.0 & 12.62 $\pm$0.1 & 10.49 $\pm$0.1  & 9.51 $\pm$0.2  & \textbf{9.23 $\pm$0.2} \\

  \hline \hline

\end{tabular}

\end{center}
\end{table*}

In practice, we opt to utilize a corruption process that does not hurt hidden abstracts significantly. This is particularly important since SAD is sensitive to large variations. For this purpose, isotropic Gaussian noise ($\mathbf{\tilde{x}} | \mathbf{x} \sim \text{N}(\mathbf{x} | 0, \sigma^2)$) and mask noise (i.e. it randomly chooses a portion of the elements) are jointly used to perturb the data.  Additionally, we limit the mask noise to alter at most $40\%$ of the original elements with the additive noise. However, the value of this parameter can be reduced for noisy data~\cite{holzwarth2003hysens}. Fig.~\ref{fig:corr} visualizes the original spectral signatures and their corrupted versions under this assumption.

\subsection{Fractional Abundance  Estimation}

After the endmember extraction, we need to find the fractional abundances of each material on data pixels. In our proposed method, we can obtain these values in two different ways.

\begin{itemize}
\item We can use the hidden abstracts $\mathbf{y}$ of our model for each sample. 

\item Also, we can solve an inverse problem with the estimated endmembers $\mathbf{W}^{(d)}$ since the decoder layer is linear. 
\end{itemize}

As explained in~\cite{dmitry2016}, due to the averaging operations for mean and variance values in batch normalization, hidden abstracts tend to generate different responses in test time and this can affect the accuracy of estimates (i.e., depending on the distribution of samples per material in the scene.). Therefore, we adapt the Simplex Projection Unmixing (SPU)~\cite{heylen2015nonlinear, heylen2011fully} algorithm with SAD kernel for abundance estimation. This algorithm estimates the abundances by using the mutual distances of material spectral signatures and data samples. It also assumes that nonnegativity and sum-to-one constraints are preserved in the nonlinear solution. 

In particular, the estimation of the abundances by solving an inverse problem with $\mathbf{W}^{(d)}$ and SPU empirically introduces further improvements to the performance, even though estimated hidden abstracts $\mathbf{y}$ still yield compatible results. Thus, the results for the fractional abundances are reported by the combination of these methods for our proposed method.

Finally, the proposed method should be considered as a spectral unmixing method as well as endmember extractor, since it optimizes the loss function jointly. We conduct additional experiments to analyze the abundance performance and a discussion for these comparisons can be found in Section \RNum{3}. C.

\section{Experiments}

Here, we demonstrate the performance of the proposed method on several datasets. For this purpose, we report quantitative and qualitative results of our proposed method on the hyperspectral unmixing problem. The estimated spectral signatures and the fractional abundances are compared with the corresponding ground truth.

In the beginning of this section, we will summarize the details of publicly available datasets, baseline methods and evaluation metrics that we used in the experiments. Lastly, we summarize the parameter settings determined in the experiments.

\begin{figure}[h]
\centering

\subfigure{
\label{fig:a}
\includegraphics[scale=0.17]{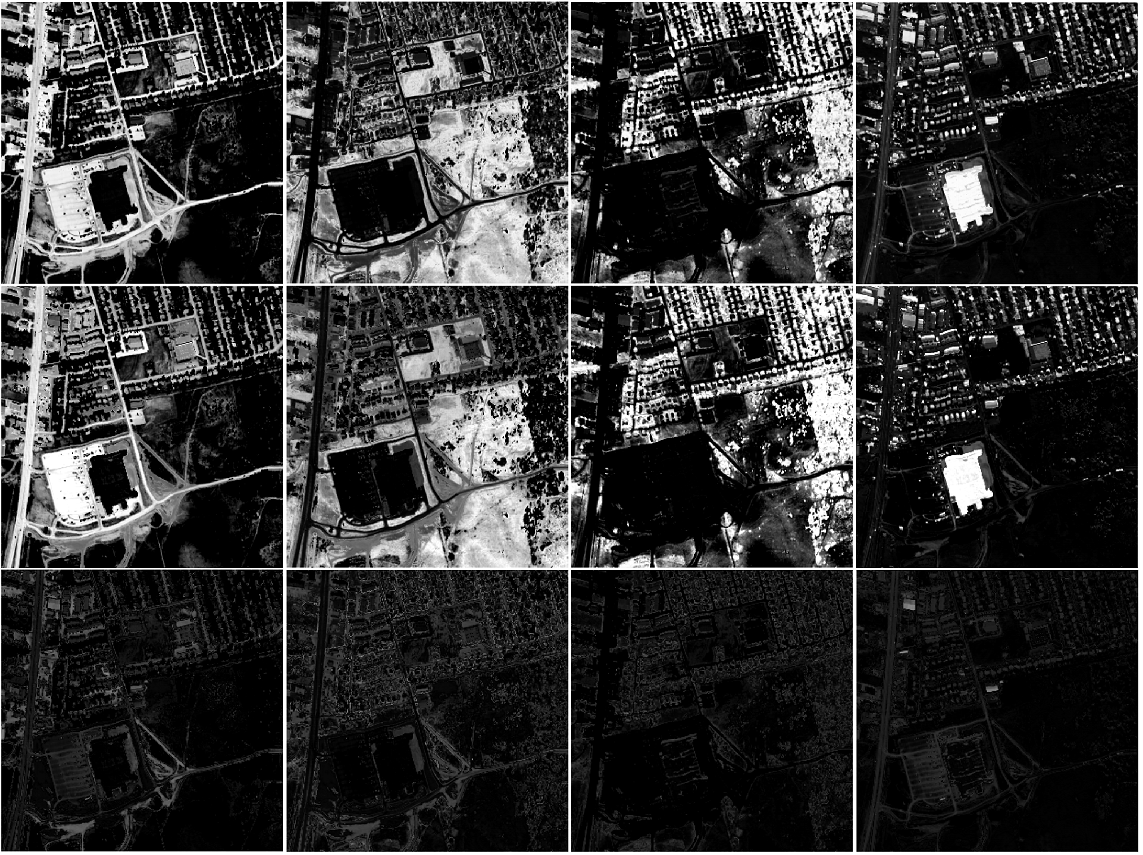}}

\caption{Fractional abundances on Urban dataset with EndNet-DMaxD method. Each column corresponds to four materials: Asphalt, Grass, Tree and Roof respectively. Also, each row indicates estimated abundance results, ground truth and their absolute differences.}

\label{fig:urban}
\end{figure}

\begin{table*}[t]
\label{tab:load}
\begin{center}

\caption{SAD and RMSE results on Samson dataset. Mean and standard deviation are reported. Best results are illustrated in bold. }
\label{table:samson}

\begin{tabular}{P{0.8cm} || P{1.45cm} P{1.45cm} P{1.45cm} P{1.45cm} P{1.45cm}  P{1.45cm}  P{1.45cm} P{1.45cm} P{1.45cm} }
  \hline
   \multirow{2}{*}{Endm.} &  \multicolumn{9}{c}{Spectral Angle Distance (SAD) ($\times 10^{-2}$)} \\ \cline{2-10}
      &  VCA  & DMaxD &  MVSA & SPICE & SCM & $l_{1|2}$-NMF & DgS-NMF & EndNet-VCA  & EndNet-DMaxD \\ \hline
     \#1  &  22.33 $\pm$2.7  &  4.04 $\pm$0.0 &  4.19 $\pm$0.0 &  45.37 $\pm$$>$9 &  1.52 $\pm$0.0 &  6.21 $\pm$7.3 & 5.64 $\pm$7.4 & 1.32 $\pm$0.2 & \textbf{1.29 $\pm$0.1}  \\
     \#2  &  4.90 $\pm$0.2  &  \textbf{2.19 $\pm$0.0}  &  5.61 $\pm$0.0 &  12.19 $\pm$$>$9 &  3.79 $\pm$0.0 & 5.23 $\pm$0.3 & 4.80 $\pm$0.3 & 4.72 $\pm$0.2  & 4.69 $\pm$0.1 \\
     \#3  &  12.29 $\pm$0.0  &  13.04 $\pm$0.0 &  18.15 $\pm$0.0 &  9.24 $\pm$$>$9 &  20.21 $\pm$0.0 & 11.97 $\pm$2.1 &  4.7 $\pm$0.3  & 3.36 $\pm$0.2  & \textbf{2.95 $\pm$0.3} \\ \hline
     Avg.  &  13.17 $\pm$1.0  &  6.42 $\pm$0.0  &  9.31 $\pm$0.0 &  22.26 $\pm$$>$9 &  8.51 $\pm$0.0 & 7.80 $\pm$3.2 & 5.05 $\pm$2.7  & 3.13 $\pm$0.2  & \textbf{2.98 $\pm$0.2} \\
  \hline \hline

   \multirow{2}{*}{} &  \multicolumn{9}{c}{Root Mean Square Error (RMSE) ($\times 10^{-2}$)} \\ \cline{2-10}
      &  VCA  & DMaxD &  MVSA & SPICE & SCM & $l_{1|2}$-NMF & DgS-NMF & EndNet-VCA  & EndNet-DMaxD \\ \hline
     \#1  &  16.92 $\pm$6.3  &  15.08 $\pm$0.0 &  19.66 $\pm$0.0 &  27.76 $\pm$$>$9 &  18.65 $\pm$0.0 & 8.58 $\pm$3.3  & 7.77 $\pm$3.8 & 5.86 $\pm$0.0 & \textbf{5.72 $\pm$0.0} \\
     \#2  &  16.64 $\pm$3.3  &  22.08 $\pm$0.0 &  23.81 $\pm$0.0 &  28.16 $\pm$2.8 &  21.65 $\pm$0.0 & 7.44 $\pm$3.7 & 7.74 $\pm$3.6 & 4.15 $\pm$0.1  & \textbf{3.84 $\pm$0.1} \\
     \#3  &  25.42 $\pm$0.6  &  26.41 $\pm$0.0 &  31.78 $\pm$0.0 &  35.57 $\pm$3.4 &  34.61 $\pm$0.0 & 5.55 $\pm$0.9  &  2.70 $\pm$0.9  & \textbf{2.07 $\pm$0.0}  & 2.11 $\pm$0.0 \\ \hline
     Avg.  &  19.66 $\pm$3.2  &  21.19 $\pm$0.0 &  25.08 $\pm$0.0 &  30.49 $\pm$5.4 &  24.97 $\pm$0.0 & 7.19 $\pm$2.4  & 6.07 $\pm$2.8  & 4.01 $\pm$0.0  & \textbf{3.88 $\pm$0.0} \\

  \hline \hline

\end{tabular}

\end{center}
\end{table*}

\subsection{Hyperspectral Datasets}

In order to make fair comparisons, we evaluate the proposed method both on synthetic data and on well-known and extensively used real datasets for endmember extraction~\cite{qian2011hyperspectral, zhu2014spectral, zhu2014structured, kruse2002comparison, holzwarth2003hysens, gader2013muufl, zare2007sparsity, zhou2016spatial, heylen2011non}. Even though synthetic data is able to measure the true fractional abundances/endmembers, it is quite hard to simulate some cases such as extreme non-linearity as in real data (even if Hapke Model or Bilinear Mixture Model (BMM) is used). In Table~\ref{table:syn}, a base experiment is performed on a synthetic dataset (i.e., we utilized Spheric dataset  from the IC Synthetic Hyperspectral Collections. To simulate the non-linear case, the noisy version with 40 db is used). It is clear that conventional models can even achieve ideal performance and the differences between the methods are quite small. Therefore, all the remaining experiments are conducted on real datasets throughout the paper to show the actual effectiveness of the methods particularly against the spectral variability. (Additional quantitative results are on the project website.) 

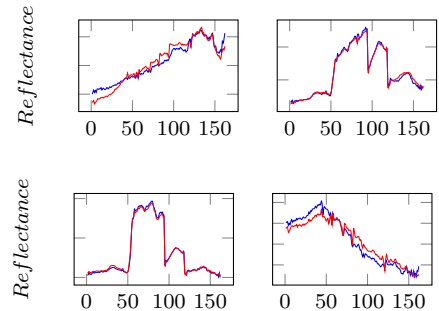
\begin{figure}[b!]
\centering
\
\subfigure{
\begin{tikzpicture}
  \begin{axis} [
      width=3.7cm, height=2.8cm,
      ylabel = {\footnotesize $Reflectance$},
      ylabel style={yshift=-0.55cm}, yticklabels={,,}
      xlabel style={yshift=0.2cm},
      xticklabel style = {font=\footnotesize},
      yticklabel style = {font=\footnotesize}
]
    \addplot [smooth,blue] table {urban_data_g1.txt};
    \addplot [smooth,red] table {urban_data_c1.txt};
  \end{axis}
\end{tikzpicture}}
\subfigure{
\begin{tikzpicture}
  \begin{axis}[
      width=3.7cm, height=2.8cm,yticklabels={,,},
      xlabel style={yshift=0.2cm},
      xticklabel style = {font=\footnotesize},
]
    \addplot [smooth,blue] table {urban_data_g2.txt};
    \addplot [smooth,red] table {urban_data_c2.txt};
  \end{axis}

\end{tikzpicture}}

\subfigure{
\begin{tikzpicture}
  \begin{axis}[
      width=3.7cm, height=2.8cm,
      ylabel = {\footnotesize $Reflectance$},
      ylabel style={yshift=-0.55cm}, yticklabels={,,}
      xlabel style={yshift=0.2cm},
      xticklabel style = {font=\footnotesize},
      yticklabel style = {font=\footnotesize}
]
    \addplot [smooth,blue] table {urban_data_g3.txt};
    \addplot [smooth,red] table {urban_data_c3.txt};
  \end{axis}

\end{tikzpicture}}
\subfigure{
\begin{tikzpicture}
  \begin{axis}[
      width=3.7cm, height=2.8cm,yticklabels={,,},
      xlabel style={yshift=0.2cm},
      xticklabel style = {font=\footnotesize},
]
    \addplot [smooth,blue] table {urban_data_g4.txt};
    \addplot [smooth,red] table {urban_data_c4.txt};
  \end{axis}

\end{tikzpicture}}

%\subfigure{
%\begin{tikzpicture}
% \begin{axis} [
%      width=3cm, height=3.8cm,
%      xlabel = \footnotesize $Bands$,
%      ylabel = {\footnotesize $Reflectance$},
%      ylabel style={yshift=-0.55cm}, yticklabels={,,}
%      xlabel style={yshift=0.2cm},
%      xticklabel style = {font=\footnotesize},
%      yticklabel style = {font=\footnotesize}
%]
%    \addplot [smooth,blue] table {urban_data_c1.txt};
%  \end{axis}
%\end{tikzpicture}}
%\subfigure{
%\begin{tikzpicture}
%  \begin{axis}[
%      width=3cm, height=3.8cm,yticklabels={,,},
%      xlabel = \small $Bands$,
%      xlabel style={yshift=0.2cm},
%      xticklabel style = {font=\footnotesize},
%]
%    \addplot [smooth,blue] table {urban_data_c2.txt};
%  \end{axis}
%
%
%\end{tikzpicture}}
%\subfigure{
%\begin{tikzpicture}
%  \begin{axis}[
%      width=3cm, height=3.8cm,yticklabels={,,},
%      xlabel = \small $Bands$,
%      xlabel style={yshift=0.2cm},
%      xticklabel style = {font=\footnotesize},
%]
%    \addplot [smooth,blue] table {figures/urban/data_c3.txt};
%  \end{axis}
%
%\end{tikzpicture}}
%\subfigure{
%\begin{tikzpicture}
%  \begin{axis}[
%      width=3cm, height=3.8cm,yticklabels={,,},
%      xlabel = \small $Bands$,
%      xlabel style={yshift=0.2cm},
%      xticklabel style = {font=\footnotesize},
%]
%    \addplot [smooth,blue] table {figures/urban/data_c4.txt};
%  \end{axis}
%
%\end{tikzpicture}}

\caption{Endmember signatures (Asphalt, Grass, Tree and Roof) on Urban dataset for ground truth (blue) and EndNet-DMaxD (red).}

\label{fig:endurban}
\end{figure}

\noindent
\textbf{Urban~\cite{qian2011hyperspectral, zhu2014spectral}:} The pixel resolution of data is 307  $\times$ 307. Several channels (1-4, 76, 87, 101-111, 136-153 and 198-210) are removed due to water-vapor absorption and atmospheric effects. There are four constituent materials: Asphalt ($\#1$), Grass ($\#2$), Tree ($\#3$), Roof ($\#4$).

\noindent
\textbf{Samson~\cite{zhu2014spectral}:} Data is generated by the SAMSON sensor. It contains 156 channels and covers the spectral range from 0.4 to 0.9$\mu$m. In order to make fair comparisons, a subimage of the original data (95 $\times$ 95) is considered as in~\cite{zhu2014spectral}. Three material types are observed in the scene: Soil ($\#1$), Tree ($\#2$) and Water ($\#3$).

\noindent
\textbf{Jasper Ridge~\cite{zhu2014structured}:} Data is captured by the AVIRIS sensor. A subimage of 100 $\times$ 100 pixels of the original data is used. Some of the channels (1-3, 108-112, 154-166 and 220-224) are discarded due to the atmospheric effects and water-vapor absorption. Tree ($\#1$), Water ($\#2$), Soil ($\#3$) and Road ($\#4$) are observed in the scene.

\begin{table}[b]
\label{tab:comp}

\begin{center}

\caption{SAD Results on Synthetic Spheric dataset (40 db noise).}
\label{table:syn}

\begin{tabular}{P{0.6cm} || P{1.5cm} | P{1.5cm} |  P{2.2cm}}
  \hline
  \multicolumn{4}{c}{Spectral Angle Distance (SAD) ($\times 10^{-2}$)}  \\ \cline{1-4}
  \hline
       &  VCA &   DMaxD     &  EndNet-DMaxD \\
  \hline
       \#1   &  0.59 & 3.91 & 1.29 \\ 
       \#2   &  0.03 & 0.60 & 0.43 \\ 
       \#3   &  0.11 & 2.12 & 1.38  \\ 
       \#4   &  0.10 & 1.00 & 1.16  \\ 
       \#5   &  0.14 & 0.84 & 0.61  \\ 
     \hline  
      Avg.  &  0.19 & 1.69 & 0.97  \\
  \hline \hline

\end{tabular}
\end{center}

\end{table}

\noindent
\textbf{Cuprite~\cite{qian2011hyperspectral, zhu2014spectral, kruse2002comparison}:} Data is captured with the AVIRIS over Cuprite, Nevada. It has 188 spectral reflectance bands covering the wavelength range from 0.4 to 2.5 $\mu$m. It has a nominal ground resolution of 20 m and a spectral resolution of 10 nm. A subimage of the original data~\cite{zhu2014spectral} (250 $\times$ 190 pixels) is considered. Noisy  (1-2 and 221-224) and water-vapor absorption (104-113 and 148-167) channels are also removed from data. This dataset hosts 12 unique mineral spectral signatures: Alunite  ($\#1$), Andradite ($\#2$), Buddingtonite ($\#3$), Dumortierite ($\#4$), Kaolinite$_1$ ($\#5$), Kaolinite$_2$ ($\#6$), Muscovite ($\#7$), Montmorillonite ($\#8$), Nontronite ($\#9$), Pyrope ($\#10$), Sphene ($\#11$) and Chalcedony ($\#12$).

\noindent
\textbf{University of Pavia~\cite{holzwarth2003hysens}:} Data is recorded by the ROSIS sensor over Pavia, Italy. The spectral band number is 103 and the spectral range is varied from 0.43 to 0.86 $\mu$m. The spatial pixel resolution is 610 $\times$ 340 and the ground resolution is approximately 1.3 m. It comprises 9 labeled classes that covers different man-made structures and natural objects. Since bricks-gravel and asphalt-bitumen have similar spectral signatures, we considered these classes as joint classes~\cite{zhou2016spatial}. In the experiment, 7 labeled classes are used: Asphalt-Bitumen  ($\#1$), Meadows ($\#2$), Trees  ($\#3$), Metal Sheets  ($\#4$), Bare Soil  ($\#5$), Gravel-Bricks ($\#6$) and Shadow ($\#7$).

\begin{figure}
\centering

\subfigure{
\label{fig:a}
\includegraphics[scale=0.11]{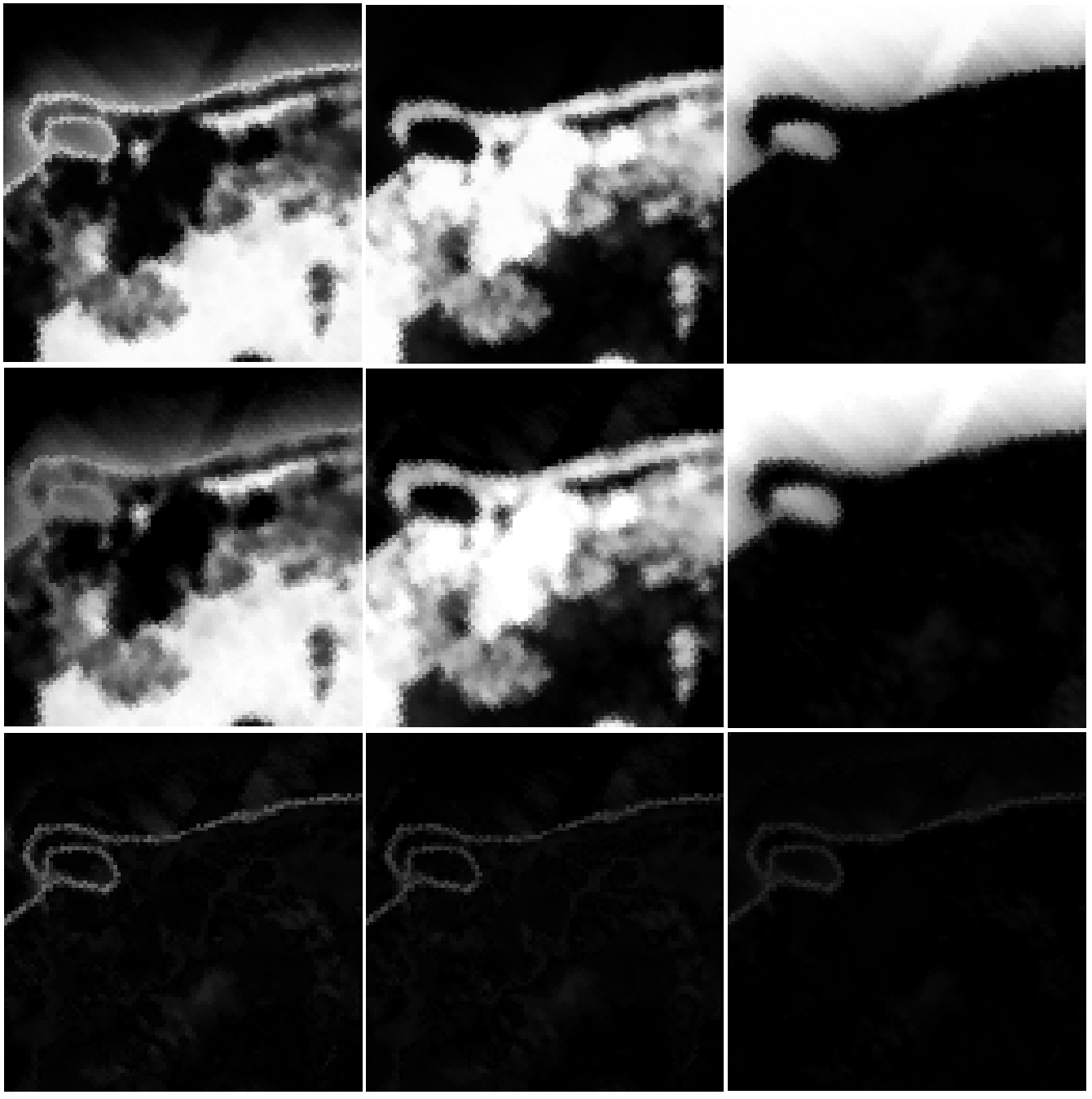}}

\caption{Fractional abundances on Samson dataset with EndNet-DMaxD method. Results for three materials, Soil, Tree and Water, are reported in different columns. Estimated fractional abundances, ground truth and theri absolute differences are illustrated in columns respectively.}

\label{fig:samson}
\end{figure}

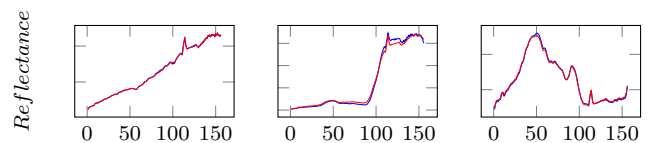
\begin{figure}[b!]
\centering

\subfigure{
\begin{tikzpicture}
  \begin{axis} [
      width=3.7cm, height=2.8cm,
      ylabel = {\footnotesize $Reflectance$},
      ylabel style={yshift=-0.55cm}, yticklabels={,,}
      xlabel style={yshift=0.2cm},
      xticklabel style = {font=\footnotesize},
      yticklabel style = {font=\footnotesize}
]
    \addplot [smooth,blue] table {samson_data_g1.txt};
    \addplot [smooth,red] table {samson_data_c2.txt};
  \end{axis}
\end{tikzpicture}}
\subfigure{
\begin{tikzpicture}
  \begin{axis}[
      width=3.7cm, height=2.8cm,yticklabels={,,},
      xlabel style={yshift=0.2cm},
      xticklabel style = {font=\footnotesize},
]
    \addplot [smooth,blue] table {samson_data_g2.txt};
    \addplot [smooth,red] table {samson_data_c3.txt};
  \end{axis}

\end{tikzpicture}}
\subfigure{
\begin{tikzpicture}
  \begin{axis}[
      width=3.7cm, height=2.8cm,yticklabels={,,},
      xlabel style={yshift=0.2cm},
      xticklabel style = {font=\footnotesize},
]
    \addplot [smooth,blue] table {samson_data_g3.txt};
    \addplot [smooth,red] table {samson_data_c1.txt};
  \end{axis}

\end{tikzpicture}}

%\subfigure{
%\begin{tikzpicture}
%  \begin{axis} [
%      width=3.7cm, height=2.8cm,
%      xlabel = \footnotesize $Bands$,
%      ylabel = {\footnotesize $Reflectance$},
%      ylabel style={yshift=-0.55cm}, yticklabels={,,}
%      xlabel style={yshift=0.2cm},
%      xticklabel style = {font=\footnotesize},
%      yticklabel style = {font=\footnotesize}
%]
%    \addplot [smooth,blue] table {samson_data_c2.txt};
%  \end{axis}
%\end{tikzpicture}}
%\subfigure{
%\begin{tikzpicture}
%  \begin{axis}[
%      width=3.7cm, height=2.8cm,yticklabels={,,},
%      xlabel = \small $Bands$,
%      xlabel style={yshift=0.2cm},
%      xticklabel style = {font=\footnotesize},
%]
%    \addplot [smooth,blue] table {samson_data_c3.txt};
%  \end{axis}
%
%\end{tikzpicture}}
%\subfigure{
%\begin{tikzpicture}
%  \begin{axis}[
%      width=3.7cm, height=2.8cm,yticklabels={,,},
%      xlabel = \small $Bands$,
%      xlabel style={yshift=0.2cm},
%      xticklabel style = {font=\footnotesize},
%]
%    \addplot [smooth,blue] table {samson_data_c1.txt};
%  \end{axis}
%
%\end{tikzpicture}}

\caption{Endmember signatures (Soil, Tree and Water) on Samson dataset for ground truth (blue) and EndNet-DMaxD (red).}

\label{fig:endsamson}
\end{figure}

\noindent
\textbf{Mississippi Gulfport~\cite{gader2013muufl}:} Data is collected from the University of Southern Mississippi’s-Gulfpark Campus. It has 64 spectral reflectance bands covering the wavelength range from 0.368 to 1.043 $\mu$m. The spatial resolution of the data is  1 m. It contains 11 man-made structures and natural materials. We eliminated sidewalk, yellow curb, cloth panels and water classes due to the lack of sufficient amounts of data samples. The remaining classes are used in the experiments:  Trees  ($\#1$), Grass Pure  ($\#2$), Grass Ground  ($\#3$), Dirt $\&$ Sand  ($\#4$), Road  ($\#5$), Shadow($\#6$) and Building ($\#7$).

\begin{table*}[t]
\label{tab:load}
\begin{center}

\caption{SAD and RMSE results on Jasper Ridge dataset. Mean and standard deviation are reported. Best results are illustrated in bold. }
\label{table:jasper}

\begin{tabular}{P{0.8cm} || P{1.45cm} P{1.45cm} P{1.45cm} P{1.45cm} P{1.45cm}  P{1.45cm}  P{1.45cm} P{1.45cm} P{1.45cm} }
  \hline
   \multirow{2}{*}{Endm.} &  \multicolumn{9}{c}{Spectral Angle Distance (SAD) ($\times 10^{-2}$)} \\ \cline{2-10}
      &  VCA  & DMaxD &  MVSA & SPICE & SCM & $l_{1|2}$-NMF & DgS-NMF & EndNet-VCA  & EndNet-DMaxD \\ \hline
     \#1   &  24.34 $\pm$5.3 &  15.58 $\pm$0.0   &  18.41 $\pm$1.2   &  40.95 $\pm$$>$9 &  7.34 $\pm$0.0 &  15.10 $\pm$0.3 &  \textbf{4.66 $\pm$0.2}  &  8.23 $\pm$3.1  &  4.99 $\pm$0.4  \\
     \#2   &  25.21 $\pm$0.4 &  25.39 $\pm$0.0   &  $>$99 $\pm$1.0   &  59.83 $\pm$$>$9 &  7.80 $\pm$0.0 & 4.60 $\pm$0.0 &  4.60 $\pm$0.0  &  \textbf{3.62 $\pm$0.4}   &  4.23 $\pm$0.9  \\
     \#3   &  34.81 $\pm$$>$9 &  13.35 $\pm$0.0   &  26.27 $\pm$0.7   &  28.03 $\pm$$>$9 &  5.61 $\pm$0.0 & 6.16 $\pm$0.5 &   5.66 $\pm$0.2   &  6.23 $\pm$0.7  &  \textbf{4.47 $\pm$0.3}  \\
     \#4   &  50.56 $\pm$9.3 &  10.69 $\pm$0.0   &  7.01 $\pm$0.8   &  86.19 $\pm$$>$9 &  3.36 $\pm$0.0 & 9.81 $\pm$0.1 &  6.73 $\pm$0.1 &  67.06 $\pm$0.5  &  \textbf{1.96 $\pm$0.2}  \\ \hline
     Avg.  &  33.73 $\pm$6.2 &  16.25 $\pm$0.0   &  37.92 $\pm$0.9   &  53.75 $\pm$$>$9 &  6.02 $\pm$0.0 & 7.19 $\pm$2.4 &  5.41 $\pm$0.1   &  21.29 $\pm$1.2   &  \textbf{3.91 $\pm$0.5}  \\
  \hline \hline

   \multirow{2}{*}{} &  \multicolumn{9}{c}{Root Mean Square Error (RMSE) ($\times 10^{-2}$)} \\ \cline{2-10}
      &  VCA  & DMaxD &  MVSA & SPICE & SCM & $l_{1|2}$-NMF & DgS-NMF & EndNet-VCA  & EndNet-DMaxD \\ \hline
     \#1   &  11.99 $\pm$5.2 &  15.14 $\pm$0.0   &  20.80 $\pm$0.4   &  22.05 $\pm$6.9 &  10.31 $\pm$0.0 &  16.16 $\pm$0.5 &  11.66 $\pm$0.2  &  8.19 $\pm$2.6  &  \textbf{8.24 $\pm$0.4}  \\
     \#2   &  11.59 $\pm$0.9 &  21.57 $\pm$0.0   &  24.03 $\pm$0.7   &  27.08 $\pm$$>$9 &  14.51 $\pm$0.0 & 5.57 $\pm$0.0 &  \textbf{4.13 $\pm$0.0}  &  26.18 $\pm$0.5   &  6.17 $\pm$0.3  \\
     \#3   &  12.54 $\pm$6.9 &  13.89 $\pm$0.0   &  20.30 $\pm$1.8   &  21.35 $\pm$5.8 &  9.70 $\pm$0.0 & 17.02 $\pm$0.4 &   11.13 $\pm$0.3   &  19.91 $\pm$0.9   &  \textbf{8.98 $\pm$0.2}  \\
     \#4   &  14.46 $\pm$3.2 &  15.76 $\pm$0.0   &  12.87 $\pm$0.2   &  27.44 $\pm$$>$9 &  7.87 $\pm$0.0 & 6.73 $\pm$0.2 & \textbf{5.68 $\pm$0.1} &  30.65 $\pm$0.3  &  8.55 $\pm$0.1  \\ \hline
     Avg.  &  12.65 $\pm$4.1 &  16.59 $\pm$0.0   &  19.50 $\pm$0.8   &  24.48 $\pm$8.1 &  10.60 $\pm$0.0 & 11.37 $\pm$0.2 &  8.15 $\pm$0.2   &  21.22 $\pm$1.1   &  \textbf{7.96 $\pm$0.3}  \\

  \hline \hline

\end{tabular}

\end{center}
\end{table*}

\subsection{Baselines, Metrics and Parameter Settings}

\noindent
\textbf{Baselines:} We compare our proposed method, EndNet, with several open-source hyperspectral unmixing algorithms throughout this study: 
\begin{itemize}
  \item Vertex Component Analysis (VCA)~\cite{nascimento2005vertex} (the code is available on \url{http://www.lx.it.pt/bioucas/code.htm})
  \item Minimum Volume Simplex Analysis (MVSA)~\cite{li2008minimum} (the code is available on \url{http://www.lx.it.pt/bioucas/code.htm})
  \item Sparsity Promoting Iterated Constrained Endmembers (SPICE)~\cite{zare2007sparsity} (the code is available on \url{https://github.com/GatorSense/SPICE})
  \item Spatial Compositional Model (SCM)~\cite{zhou2016spatial} (the code is available on \url{https://github.com/zhouyuanzxcv/Hyperspectral})
  \item Distance-MaxD (DMaxD)~\cite{heylen2011non} (the code is available on \url{https://sites.google.com/site/robheylenresearch/code})
\end{itemize}

\begin{figure}
\centering
\subfigure{
\label{fig:a}
\includegraphics[scale=0.11]{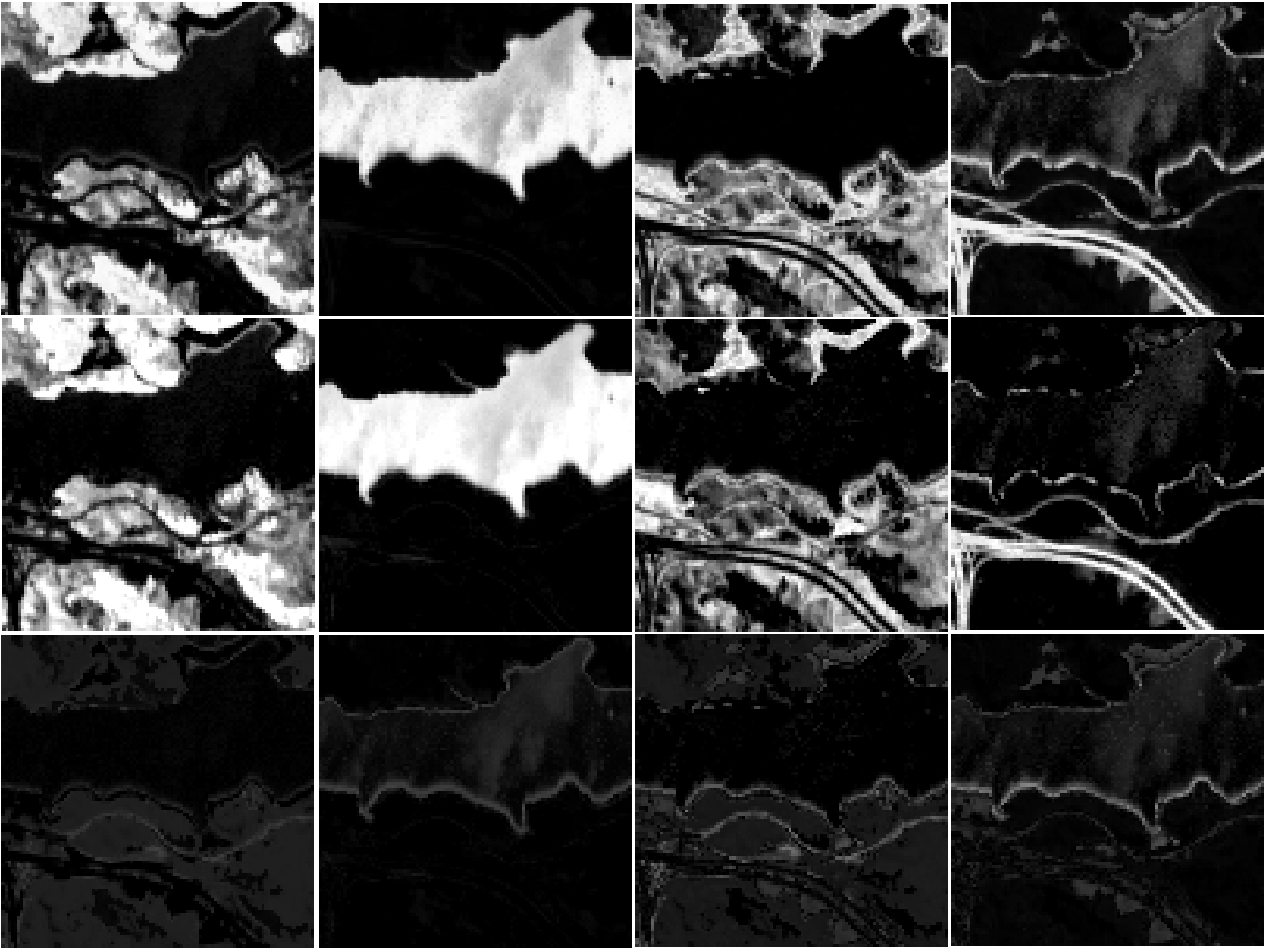}}

\caption{Estimated fractional abundances and the absolute differences with the ground truth on Jasper Ridge dataset for EndNet-DMaxD method. Four materials, Tree, Water, Soil and Road, are illustrated in different columns.}

\label{fig:jasper}
\end{figure}

\begin{figure}[b!]
\centering

\subfigure{
\begin{tikzpicture}
  \begin{axis} [
      width=3.7cm, height=2.8cm,
      ylabel = {\footnotesize $Reflectance$},
      ylabel style={yshift=-0.55cm}, yticklabels={,,}
      xlabel style={yshift=0.2cm},
      xticklabel style = {font=\footnotesize},
      yticklabel style = {font=\footnotesize}
]
    \addplot [smooth,blue] table {jasper_data_g1.txt};
    \addplot [smooth,red] table {jasper_data_c4.txt};
  \end{axis}
\end{tikzpicture}}
\subfigure{
\begin{tikzpicture}
  \begin{axis}[
      width=3.7cm, height=2.8cm,yticklabels={,,},
      xlabel style={yshift=0.2cm},
      xticklabel style = {font=\footnotesize},
]
    \addplot [smooth,blue] table {jasper_data_g2.txt};
    \addplot [smooth,red] table {jasper_data_c1.txt};
  \end{axis}

\end{tikzpicture}}

\subfigure{
\begin{tikzpicture}
  \begin{axis}[
      width=3.7cm, height=2.8cm,
      ylabel = {\footnotesize $Reflectance$},
      ylabel style={yshift=-0.55cm}, yticklabels={,,}
      xlabel style={yshift=0.2cm},
      xticklabel style = {font=\footnotesize},
      yticklabel style = {font=\footnotesize}
]
    \addplot [smooth,blue] table {jasper_data_g3.txt};
    \addplot [smooth,red] table {jasper_data_c3.txt};
  \end{axis}

\end{tikzpicture}}
\subfigure{
\begin{tikzpicture}
  \begin{axis}[
      width=3.7cm, height=2.8cm,yticklabels={,,},
      xlabel style={yshift=0.2cm},
      xticklabel style = {font=\footnotesize},
]
    \addplot [smooth,blue] table {jasper_data_g4.txt};
    \addplot [smooth,red] table {jasper_data_c2.txt};
  \end{axis}

\end{tikzpicture}}

%\subfigure{
%\begin{tikzpicture}
%  \begin{axis} [
%      width=2.8cm, height=2.8cm,
%      xlabel = \footnotesize $Bands$,
%      ylabel = {\footnotesize $Reflectance$},
%      ylabel style={yshift=-0.55cm}, yticklabels={,,}
%      xlabel style={yshift=0.2cm},
%      xticklabel style = {font=\footnotesize},
%      yticklabel style = {font=\footnotesize}
%]
%    \addplot [smooth,blue] table {figures/jasper/data_c4.txt};
%  \end{axis}
%\end{tikzpicture}}
%\subfigure{
%\begin{tikzpicture}
%  \begin{axis}[
%      width=2.8cm, height=2.8cm,yticklabels={,,},
%      xlabel = \small $Bands$,
%      xlabel style={yshift=0.2cm},
%      xticklabel style = {font=\footnotesize},
%]
%    \addplot [smooth,blue] table {figures/jasper/data_c1.txt};
%  \end{axis}
%
%\end{tikzpicture}}
%\subfigure{
%\begin{tikzpicture}
%  \begin{axis}[
%      width=2.8cm, height=2.8cm,yticklabels={,,},
%      xlabel = \small $Bands$,
%      xlabel style={yshift=0.2cm},
%      xticklabel style = {font=\footnotesize},
%]
%    \addplot [smooth,blue] table {figures/jasper/data_c3.txt};
%  \end{axis}
%
%\end{tikzpicture}}
%\subfigure{
%\begin{tikzpicture}
%  \begin{axis}[
%      width=2.8cm, height=2.8cm,yticklabels={,,},
%      xlabel = \small $Bands$,
%      xlabel style={yshift=0.2cm},
%      xticklabel style = {font=\footnotesize},
%]
%    \addplot [smooth,blue] table {figures/jasper/data_c2.txt};
%  \end{axis}
%
%\end{tikzpicture}}

\caption{Endmember signatures (Tree, Water, Soil and Road) on Jasper dataset for ground truth (blue) and EndNet-DMaxD (red).}

\label{fig:endjasper}
\end{figure}
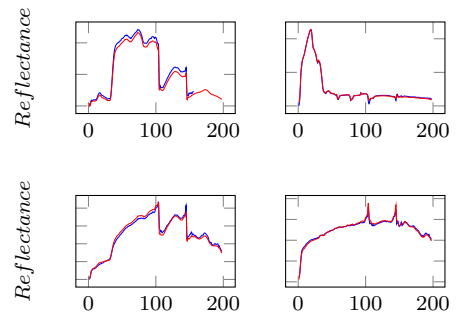

In addition to these open-source algorithms, we report the performance of recent methods for several datasets by referring directly to their reported scores, since their codes are not available: $l_{1|2}$-NMF~\cite{qian2011hyperspectral, zhu2014spectral}, DgS-NMF~\cite{zhu2014spectral}. Particularly, we should note that we compare our proposed method with the method like DgS-NMF~\cite{zhu2014spectral} to show that our proposed method outperforms even the methods that exploit spatial priors in their solutions. 

\begin{figure}
\centering

\subfigure{
\label{fig:a}
\includegraphics[scale=0.28]{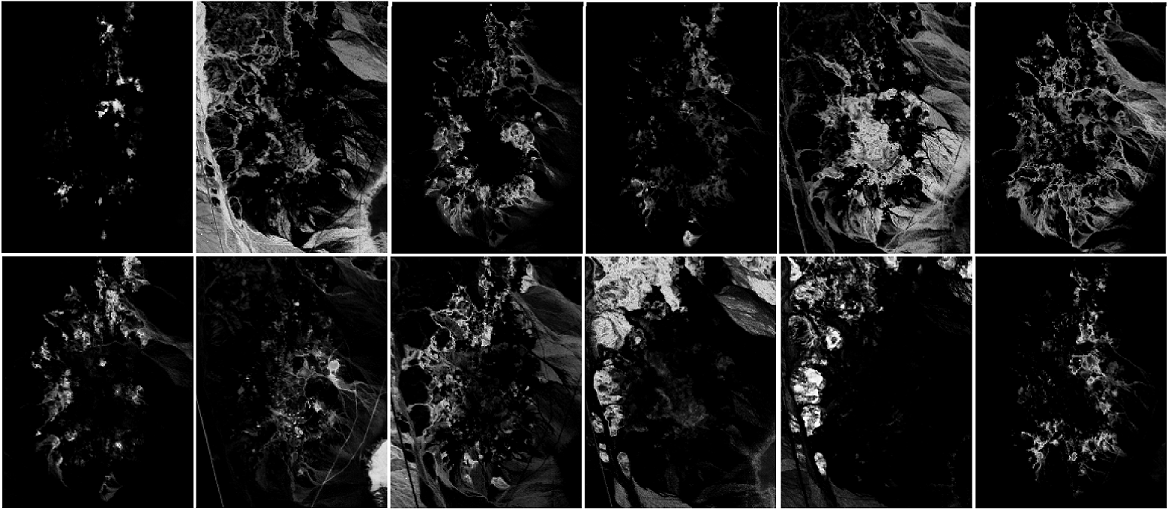}}

\caption{Estimated fractional abundance results on Cuprite dataset with the EndNet-DMaxD method. First Row: Alunite, Andradite, Buddingtonite, Dumortierite, Kaolinite$_1$, Kaolinite$_2$.  Second Row: Muscovite, Montmorillonite, Nontronite, Pyrope, Sphene, Chalcedony. }

\label{fig:cuprite}
\end{figure}

\noindent
\textbf{Metrics:} To evaluate unmixing performance and compare with ground truth, we utilize two metrics: Spectral Angle Distance (SAD) and Root Mean Square Error (RMSE):
\begin{eqnarray}
\label{eqn:bc4}
\text{SAD}(\mathbf{e}, \mathbf{\hat e}) = \text{cos}^{-1}\bigg(  \frac{\mathbf{e} \hspace{1mm} \mathbf{\hat e}}{ \| \mathbf{e} \|_2 \| \mathbf{\hat e} \|_2 } \bigg),
\end{eqnarray}

\begin{table*}[t]
\label{tab:load}
\begin{center}

\caption{SAD results on Cuprite dataset. Mean and standard deviation are reported. Best results are illustrated in bold. }
\label{table:cuprite}

\begin{tabular}{P{0.8cm} || P{1.45cm} P{1.45cm} P{1.45cm} P{1.45cm} P{1.45cm}  P{1.45cm}  P{1.45cm} P{1.45cm} P{1.45cm} }
  \hline
   \multirow{2}{*}{Endm.} &  \multicolumn{9}{c}{Spectral Angle Distance (SAD) ($\times 10^{-2}$)} \\ \cline{2-10}
     &  VCA  & DMaxD &  MVSA & SPICE & SCM & $l_{1|2}$-NMF & DgS-NMF & EndNet-VCA  & EndNet-DMaxD \\ \hline
     \#1  &  12.89 $\pm$5.3  &  \textbf{8.58 $\pm$0.0} &  23.60 $\pm$5.7 &  14.71 $\pm$1.6 &  24.43 $\pm$4.6 &  16.22 $\pm$2.0 &  12.48 $\pm$1.8 & 13.60 $\pm$2.9 & 14.18 $\pm$2.5  \\
     \#2  &  8.21 $\pm$1.9  &  \textbf{7.15 $\pm$0.0} &  $>$99 $\pm$$>$9 &  9.32 $\pm$1.6 &  7.19 $\pm$0.2 &  10.15 $\pm$3.0 &  7.59 $\pm$1.2 & 7.65 $\pm$0.1 & 7.34 $\pm$0.2 \\
     \#3  &  \textbf{9.04 $\pm$2.0}  &  11.58 $\pm$0.0 &  $>$99 $\pm$$>$9 &  10.26 $\pm$0.3 &  13.25 $\pm$0.3 &  12.50 $\pm$7.5 &  10.81 $\pm$3.2 & 11.44 $\pm$0.8 & 11.66 $\pm$1.1 \\ 
     \#4  &  9.51 $\pm$3.0  &  8.01 $\pm$0.0 &  38.36 $\pm$$>$9 &  11.91 $\pm$1.3 &  12.22 $\pm$0.9 &  13.07 $\pm$5.3 &  11.01 $\pm$2.1 & 8.38 $\pm$0.6 & \textbf{7.81 $\pm$0.7}  \\
     \#5  &  8.88 $\pm$1.5  &  \textbf{8.47 $\pm$0.0} &  25.44 $\pm$6.2 &  13.09 $\pm$2.7 &  9.20 $\pm$1.3 &  9.42 $\pm$1.8 &  9.02 $\pm$2.9 & 12.06 $\pm$0.9 & 13.16 $\pm$1.1 \\
     \#6  &  7.08 $\pm$5.2  &  8.52 $\pm$0.0 &  $>$99 $\pm$$>$9 &  9.60 $\pm$0.5 &  6.80 $\pm$0.4 &  9.82 $\pm$2.7 &  7.42 $\pm$1.2 & 6.05 $\pm$0.3 & \textbf{5.69 $\pm$0.2} \\ 
     \#7  &  17.13 $\pm$4.0  &  \textbf{10.26 $\pm$0.0} &  41.73 $\pm$$>$9 &  9.97 $\pm$0.5 &  17.15 $\pm$2.5 &  29.86 $\pm$7.4 &  20.51 $\pm$6.0 & 15.93 $\pm$1.9 & 15.45 $\pm$1.1  \\
     \#8  &  6.22 $\pm$0.4  &  6.19 $\pm$0.0 &  81.84 $\pm$$>$9 &  8.60 $\pm$0.5 &  6.28 $\pm$0.2 &  10.27 $\pm$4.7 &  7.27 $\pm$1.2 & 5.51 $\pm$0.3 & \textbf{5.75 $\pm$0.1} \\
     \#9  &  7.95 $\pm$1.0  &  \textbf{7.36 $\pm$0.0} &  65.95 $\pm$$>$9 &  11.45 $\pm$1.6 &  7.61 $\pm$0.1 &  12.80 $\pm$4.1 &  8.88 $\pm$1.7 & 8.36 $\pm$0.7 & 7.78 $\pm$0.1 \\ 
     \#10  &  11.11 $\pm$4.5  &  13.08 $\pm$0.0 &  88.65 $\pm$$>$9 &  6.99 $\pm$1.9 &  5.89 $\pm$0.1 &  8.12 $\pm$1.9 &  8.82 $\pm$5.0 & \textbf{5.87 $\pm$0.5} & 6.12 $\pm$0.4  \\
     \#11 &  8.40 $\pm$5.9  &  28.56 $\pm$0.0 &  52.02 $\pm$$>$9 & 8.99 $\pm$2.4 &  \textbf{7.11 $\pm$1.7} &  11.03 $\pm$3.3 &  8.15$\pm$2.1 & 9.03 $\pm$1.4 & 8.38 $\pm$1.2 \\
     \#12  &  10.07 $\pm$4.5  &  \textbf{8.23 $\pm$0.0} &  9.10 $\pm$1.35 &  8.63 $\pm$0.9 &  9.29 $\pm$0.2 &  13.72 $\pm$6.0 &  13.62 $\pm$6.3 & 10.56 $\pm$1.6 & 11.14 $\pm$0.9 \\ \hline
     Avg.  &  9.71 $\pm$2.9  &  10.54 $\pm$0.0  &  71.92 $\pm$$>$9 &  10.30 $\pm$1.3 &  10.54 $\pm$1.0 & 13.07 $\pm$4.1 & 10.46 $\pm$2.9  & \textbf{9.53 $\pm$1.0}  & \textbf{9.54 $\pm$0.8} \\
  \hline \hline

\end{tabular}

\end{center}
\end{table*}

As mentioned earlier, SAD is used to evaluate the quality of estimated endmember with ground truth by measuring angle distance. Similarly, to assess the accuracy of the estimated abundances, we utilized the RMSE metric as:
\begin{eqnarray}
\label{eqn:bc4}
\text{RMSE}(\mathbf{y}, \mathbf{\hat y}) = \sqrt{\frac{1}{N} \| \mathbf{y} - \mathbf{\hat y} \|^2_2}.
\end{eqnarray}

Note that smaller values indicate better performance for both metrics.

\noindent
\textbf{Parameter Settings:} We fixed the default parameter values of the baseline methods in our experiments. We tuned the parameters of SCM algorithm for each dataset based on the observations in~\cite{zhou2016spatial}. 

\begin{table}[b]
\label{tab:comp}

\begin{center}

\caption{Average computation time of the methods for three datasets.}
\label{table:comp}

\begin{tabular}{P{2.6cm} || P{1.2cm} | P{1.2cm} |  P{1.2cm}}
  \hline
  \multicolumn{4}{c}{Computation Time (Sec)}  \\ \cline{1-4}
  \hline
       &  Urban &   Samson     &  Jasper \\
  \hline
       VCA-MLM                    &  $\approx$ 3600 &  $\approx$ 349    & $\approx$ 352 \\ 
       SCM                            &  $\approx$ 1021 &  $\approx$ 187   &  $\approx$ 472 \\
       EndNet-DMaxD-SPU   &  $\approx$ 914 &    $\approx$ 789  &  $\approx$ 855 \\ 
  \hline \hline

\end{tabular}
\end{center}

\end{table}

For fractional abundance estimation, we selected Multilinear Mixing Model (MLM)~\cite{mlm2016} for MVSA, VCA and DMaxD algorithms to preserve nonlinearity (Note that we also tested these algorithms with Fully-Constrained Least Square (FCLS), Generalized Bilinear Model (GBM)~\cite{gbm2011}, Postnonlinear Mixing Model (PPNM)~\cite {super2012} and Simplex
Projection Unmixing (SPU). MLM yielded the best performance). SPICE and SCM compute the optimum abundances for the scene in their algorithms. 

For our proposed method, three parameters should be tuned by the user for different scenes, namely $p$ in the dropout layer, $\lambda_2$ in the optimization step and the percentage of mask noise. We will redefine the values of these parameters if the scene needs to be tuned by presenting reasonable explanations. Otherwise, the default values are used on each dataset ($p=1.0$, $\lambda_2=0.1$ and $40\%$ mask noise). Also, EndNet-VCA and EndNet-DMaxD in the experiments denote that which of the algorithms is used for the parameter initialization of our proposed method.

\begin{table}[b]
\label{tab:comp}

\begin{center}

\caption{Comparison of abundance estimation results for SPU (first rows) and EndNet-DMaxD (second rows).}
\label{table:spu}

\begin{tabular}{P{0.6cm} || P{2.2cm} | P{2.2cm} |  P{2.2cm}}
  \hline
  \multicolumn{4}{c}{Root Mean Square Error (RMSE) ($\times 10^{-2}$)}  \\ \cline{1-4}
  \hline
       &  Urban &   Samson     &  Jasper \\
  \hline
       \multirow{2}{*}{\#1}   &  SPU: 10.41 $\pm$0.2 &  SPU: 5.72 $\pm$0.0     & SPU: 8.24 $\pm$0.4 \\
	                                  &  EndN: 13.04 $\pm$0.3 &  EndN: 9.41 $\pm$0.1     &  EndN: 10.12 $\pm$0.6 \\ \cline{1-4}
      \multirow{2}{*}{\#2}    &  SPU: 12.24 $\pm$0.3 &  SPU: 3.84 $\pm$0.1   &  SPU: 6.17 $\pm$0.3 \\
				  &  EndN: 14.43 $\pm$0.3         &  EndN: 6.47 $\pm$0.3   &  EndN: 11.48 $\pm$0.8 \\ \cline{1-4}
       \multirow{2}{*}{\#3}   &  SPU: 8.35  $\pm$0.3 &  SPU: 2.11 $\pm$0.0   &  SPU: 8.19 $\pm$0.2 \\ 
				  &  EndN: 8.71 $\pm$0.5          &  EndN: 3.93 $\pm$0.1   &  EndN: 9.53 $\pm$0.3 \\ \cline{1-4}
      \multirow{2}{*}{\#4}    &  SPU: 5.92 $\pm$0.2 &     -                          &  SPU: 8.55 $\pm$0.1  \\
			             &  EndN: 7.59 $\pm$0.2 &  -                                      &  EndN: 12.29 $\pm$0.4  \\ \cline{1-4}
     \hline     \hline
      \multirow{2}{*}{Avg.}   &  SPU: 9.23 $\pm$0.2 &  SPU: 3.88 $\pm$0.0   &  SPU: 7.96 $\pm$0.3  \\
				  &  EndN: 10.94 $\pm$0.4 &  EndN: 5.72 $\pm$0.2   &  EndN: 10.85 $\pm$0.6  \\
  \hline \hline

\end{tabular}
\end{center}

\end{table}

Finally, the corresponding ground truth material (i.e. most similar ground truth material) is determined by measuring its highest SAD similarity score with the estimated endmembers  in the experiments.

\noindent
\textbf{Computational Complexity:} Speed and memory requirement are two important parameters for the problem. Since a stochastic gradient-based solver is utilized in the proposed method, it practically scales the problem for large-scale data~\cite{bottou2010large}. Intuitively, the computation time and memory-requirement is independent from data since batch-based learning is used. Moreover, the computation time for the proposed method on each dataset is illustrated in Table~\ref{table:comp} (all codes are implemented on Matlab). Note that the computation time of our proposed method is independent from data size and even if data size increases, the computation time will be in the same order. Lastly, the proposed method can be easily parallelized on Graphical Processing Units (GPUs) due to the neural network architecture.

\subsection{Experiments on Hyperspectral Unmixing Datasets}

In this section, we compare our proposed method with the baselines on the Urban, Samson, Jasper Ridge and Cuprite datasets. These four datasets are illustrated in Fig.~\ref{fig:data1}. For reliable assessments, tests are repeated 20 times for each method, thus mean and standard deviation of the results are reported. We also illustrate the qualitative results of the proposed method to provide visual comparisons on the estimated fractional abundances.

In Tables~\ref{table:urban}, ~\ref{table:samson} and ~\ref{table:jasper}, each row corresponds to SAD or RMSE performance of different methods for a single material. The last row at each table denotes the average performance of all materials for the corresponding metric. In Table ~\ref{table:cuprite}, only the SAD performance is reported since Cuprite dataset does not have a quantitative fractional abundance ground truth. Fig.~\ref{fig:urban},~\ref{fig:samson},~\ref{fig:jasper} and~\ref{fig:cuprite} visualize the qualitative abundance performance of our EndNet-DMaxD method on these datasets. In these figures, we provide the estimated fractional abundances, the ground truth  (if it is available) and the absolute differences of the estimated abundances. In addition, the estimated endmember and ground truth signatures are illustrated for the Urban, Samson and Jasper datasets in Fig.~\ref{fig:endurban},~\ref{fig:endsamson} and~\ref{fig:endjasper}. (The project website provides the qualitative results on different datasets.)

Lastly, note that for SCM method, the parameters are tuned to $\beta_1=10$,  $\beta_2=5$ and $\rho=0.01$ for the Samson and the Jasper datasets while they are set to $\beta_1=10$,  $\beta_2=10$ and $\rho=0.01$ for Urban and Cuprite datasets.

\noindent
\textbf{Urban:} Quantitive and qualitative results on Urban dataset are summarized in Table ~\ref{table:urban} and Fig.~\ref{fig:urban}. From the results, our proposed method achieves the best overall scores for both SAD and RMSE metrics compared to other methods. EndNet-DMaxD is approximately 4.2\% and 1.2\% better than the second best result. The second best result is obtained by DgS-NMF method that exploits spatial priors in the estimation. This is important since our proposed method outperforms the methods that use spatial information about the datasets, without exploiting any extra prior.

\begin{table*}[t]
\label{tab:load}
\begin{center}

\caption{SAD results on University of Pavia dataset. Mean and standard deviation are reported. Best results are illustrated in bold. }
\label{table:pavia}

\begin{tabular}{P{0.8cm} || P{1.45cm} P{1.45cm} P{1.45cm} P{1.45cm}  P{1.45cm}  P{1.45cm} }
  \hline
   \multirow{2}{*}{Endm.} &  \multicolumn{6}{c}{Spectral Angle Distance (SAD) ($\times 10^{-2}$)} \\ \cline{2-7}
      &  VCA  & DMaxD & SPICE & SCM & EndNet-VCA  & EndNet-DMaxD \\ \hline
      \#1   &  51.02 $\pm$8.9 &  7.50 $\pm$0.0     &  57.98 $\pm$$>$9 & 2.93 $\pm$0.0  &  \textbf{2.31 $\pm$1.1} & 2.38 $\pm$0.6    \\
      \#2   &  39.43 $\pm$$>$9 &  92.98 $\pm$0.0   &  66.33 $\pm$$>$9 & \textbf{1.85 $\pm$0.0}  &  4.19 $\pm$1.1 & 6.07 $\pm$1.8    \\
      \#3   &  28.85 $\pm$$>$9 &  $>$99 $\pm$0.0   &  12.50 $\pm$5.3 & 4.70 $\pm$0.0  &  \textbf{1.62 $\pm$0.1} & 2.03 $\pm$0.8    \\
      \#4   &  42.47 $\pm$4.6 &  30.60 $\pm$0.0   &  44.45 $\pm$$>$9 & 11.67 $\pm$0.0  &  10.11 $\pm$1.2 & \textbf{8.57 $\pm$1.7}    \\
      \#5   &  85.19 $\pm$$>$9 &  70.95 $\pm$0.0   &  $>$99 $\pm$$>$9 & 7.69 $\pm$0.0  &  \textbf{5.73 $\pm$0.6} & 7.04 $\pm$0.8    \\
      \#6   &  42.65 $\pm$8.8 &  56.93 $\pm$0.0   &  93.00 $\pm$$>$9 & 27.65 $\pm$0.0  &  4.62 $\pm$2.6 & \textbf{1.59 $\pm$0.5}    \\
      \#7   &  53.04 $\pm$3.7 &  53.06 $\pm$0.0   &  94.72 $\pm$$>$9 & \textbf{5.29 $\pm$0.0}  &  55.80 $\pm$0.3 & 13.40 $\pm$1.9    \\    \hline
     Avg.  &  48.96 $\pm$$>$9 &  61.33 $\pm$0.0   &  67.79 $\pm$$>$9 & 8.82 $\pm$0.0 &  12.07 $\pm$1.0 &  \textbf{5.87 $\pm$1.2}    \\
  \hline \hline

\end{tabular}

\end{center}
\end{table*}

Fig.~\ref{fig:urban} visualizes the estimated fractional abundance for the dataset with EndNet-DMaxD method. To compare the estimation accuracy, absolute differences of the estimated fractional abundance and ground truth are provided for each material (i.e rows). From these results, we can clearly observe that the estimation error is usually concentrated on the object boundaries. Despite the fact that our proposed method might yield false estimates where high spectral mixtures have occurred, we must emphasize that the ground truth can still exhibit some noise since the determination of perfect abundance labels of these locations can be challenging.

\begin{figure}[t]
\centering
\subfigure{
\label{fig:a}
\includegraphics[scale=0.15]{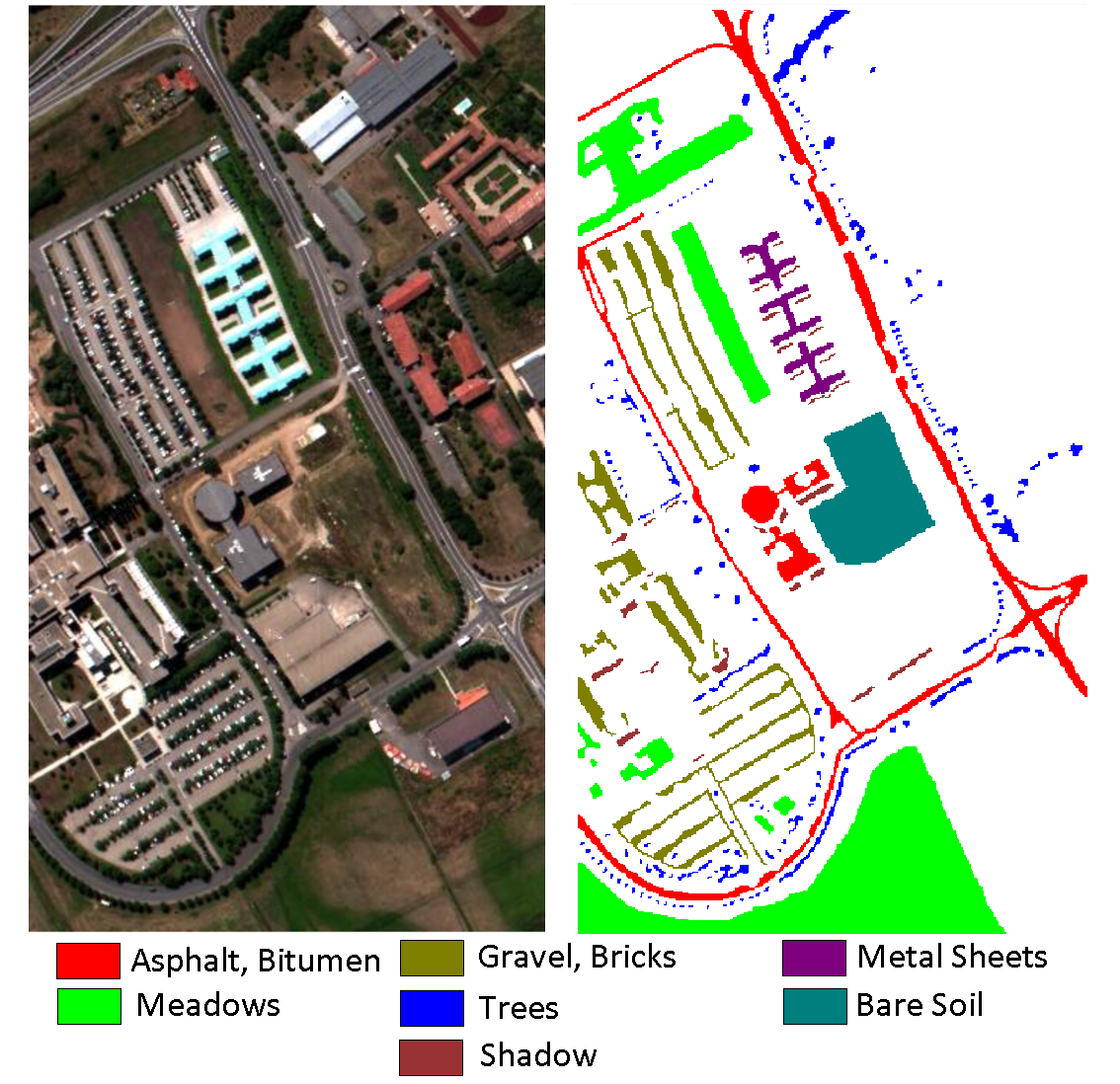}}
\subfigure{
\label{fig:a}
\includegraphics[scale=0.14]{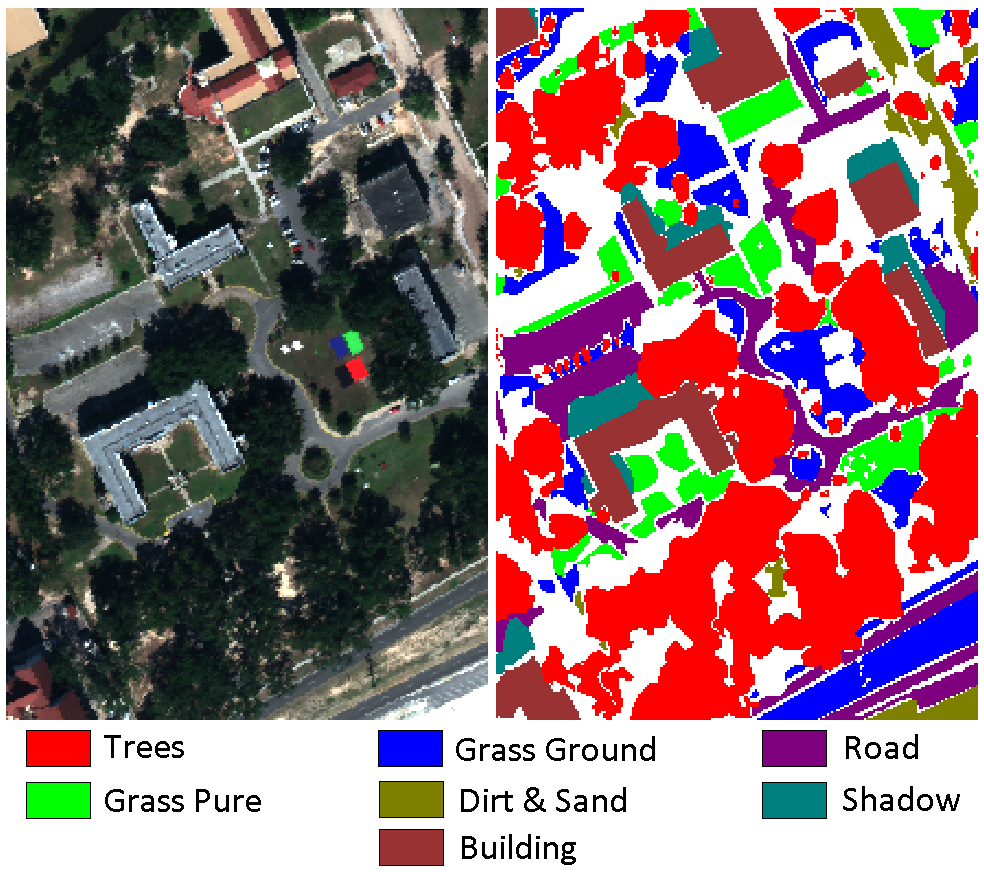}}

\caption{Two real hyperpsectral datasets with pixel-wise class labels: University of Pavia and Mississippi Gulfport respectively.}

\label{fig:data2}
\end{figure}

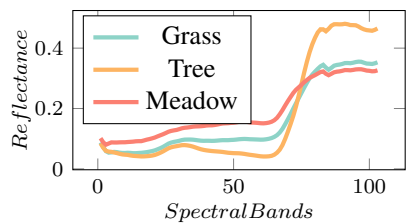
\begin{figure}[b!]
\centering

\subfigure{
\begin{tikzpicture}
  \begin{axis} [
      width=6.0cm, height=3.7cm,
      xlabel = \footnotesize $Spectral Bands$,
      ylabel = {\footnotesize $Reflectance$},
      ylabel style={yshift=-0.55cm},
      xlabel style={yshift=0.2cm},
      xticklabel style = {font=\footnotesize},
      yticklabel style = {font=\footnotesize},
      legend pos=north west
]
    \addplot [ultra thick,ryb1] table {pavia_data_gr.txt}; \addlegendentry{Grass}
    \addplot [ultra thick,ryb6] table {pavia_data_tr.txt}; \addlegendentry{Tree}
    \addplot [ultra thick,ryb4] table {pavia_data_me.txt}; \addlegendentry{Meadow}

;
  \end{axis}
\end{tikzpicture}}

\caption{Three highly correlated materials observed in University of Pavia dataset: Grass, Tree and Meadow. Even if Grass and Meadow are considered as one constituent material in the original dataset, they have two distinct spectral signatures as shown in the plot.}

\label{fig:endplus}
\end{figure}

\noindent
\textbf{Samson:} Results are shown in Table ~\ref{table:samson} and Fig.~\ref{fig:samson}. As for the previous scenarios, our proposed methods obtain the best overall performance for both metrics, and the RMSE results are prominently improved by the proposed method. Approximately $2\%$ and $3\%$ improvements are introduced for SAD and RMSE metrics respectively. From Fig.~\ref{fig:samson}, it can be seen that the absolute difference is quite small. Only the water-ground intersection areas yield a small estimation error where all three materials can contribute.

\noindent
\textbf{Jasper Ridge:} Remark that soil ($\#3$) and road ($\#4$) are highly mixtured materials in this scene. Therefore, we redefine  $\lambda_2$ as $0.4$ to increase the sparsity of the material abundances in the proposed method. Table ~\ref{table:jasper} shows the SAD and RMSE performance of the methods. From the results, our proposed method introduces small improvements over DgS-NMF. However, it should also be noted that even if the use of spatial priors provides huge advantages, our method still yields the best results for highly mixtured scenes. Fig.~\ref{fig:jasper} illustrates the absolute error with respect to the ground truth. The error is intensified at the boundaries of water-ground and the regions designated road-soil.

Lastly, an important observation for one of our proposed methods is that the performance of EndNet-VCA drastically decreases on this dataset while EndNet-DMaxD maintains the exceptional performance. Our explanation of this result is that the DMaxD method is inclined to detect more uncorrelated spectral signatures from data than VCA (even if the estimates do not need to be the true results), since it maximizes the distances of the estimated spectral signatures with one another. Also, VCA can ignore the materials that are numerically small in a scene. Thus, considering the methods such as DMaxD for the proposed method initialization can be more advantageous.

\begin{figure}[t]
\centering
\subfigure{
\label{fig:a}
\includegraphics[scale=0.31]{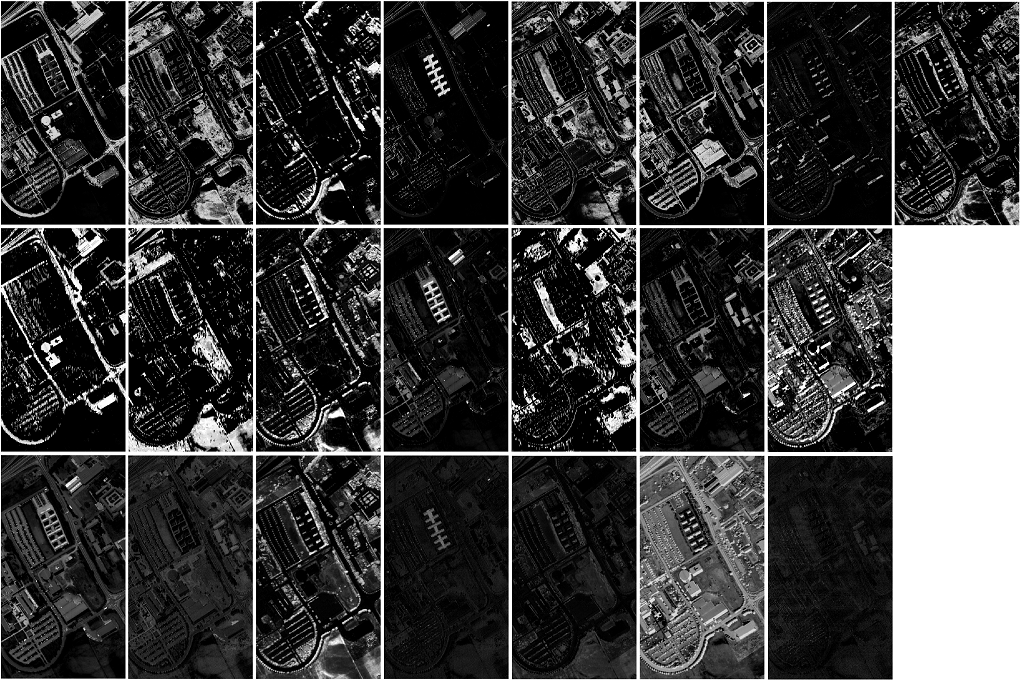}}

\caption{Estimated fractional abundances on University of Pavia dataset. Rows indicate methods: EndNet-DMaxD, SCM and SPICE, while each column corresponds to the fractional abundance of a single material: Asphalt-Bitumen  ($\#1$), Meadows ($\#2$), Trees  ($\#3$), Metal Sheets  ($\#4$), Bare Soil  ($\#5$), Gravel-Bricks ($\#6$) and Shadow ($\#7$) respectively. We also visualize the fractional abundance of Grass material obtained by EndNet-DMaxD method as the last image. }

\label{fig:pavia}
\end{figure}

\noindent
\textbf{Cuprite:} Experimental results for this dataset are reported in Table ~\ref{table:cuprite} and Fig.~\ref{fig:cuprite}. From these results, it is clear that the methods using the unmixed pixel assumption such as VCA, DMaxD etc. obtain high performance. Therefore, we decreased the value of the sparsity term $\lambda_2$ to $0.001$. Another important feature of the dataset is that it consists of quite correlated materials. For this reason, we also set $p$ to $0.8$ to improve the convergence of the parameters.

The experimental results show that the proposed methods produce the lowest SAD results compared to the baseline methods. Fig.~\ref{fig:cuprite} visualizes the estimated fractional abundance for each material with the EndNet-DMaxD method. Since this dataset has no ground truth for the true abundances, we can only make visual comparisons to analyze the consistency with other methods~\cite{kruse2002comparison}. When we compare our results with~\cite{kruse2002comparison}, we observed that the estimated fractional abundances are consisted with the reported visual results.

\noindent
\textbf{Comparison of Abundance Map Estimation:} The detailed experimental results for the abundance estimation of SPU and EndNet-DMaxD are reported in Table~\ref{table:spu}. For three datasets, the SPU method introduces further improvements to the performance compared to EndNet-DMaxD abundance estimates. However, note that EndNet-DMaxD achieves similar or better performance compared to the baseline methods.

\subsection{Experiments on Hyperspectral Classification Datasets}

In this section, we analyze SAD and the qualitative performance of the baseline methods on the University of Pavia and Mississippi Gulfport datasets. In particular, these experimental results can provide a basis for future hyperspectral classification methods which could use the proposed method in these studies. Tests are repeated 20 times. Fig.~\ref{fig:data2} illustrates these datasets and their pixel-wise class labels. Note that no RMSE performance on abundance estimates are reported in the paper on the contrary to~\cite{zhou2016spatial}, since class labels are quite noisy and this information might not convey the true performance of the method.

Due to the fact that there is no endmember ground truth available for these datasets, we used the average spectra of the pixels for each material as the ground truth in SAD comparisons as proposed in~\cite{zhou2016spatial, zare2013piecewise}.

\begin{figure}[b]
\centering

\subfigure{
\label{fig:a}
\includegraphics[scale=0.31]{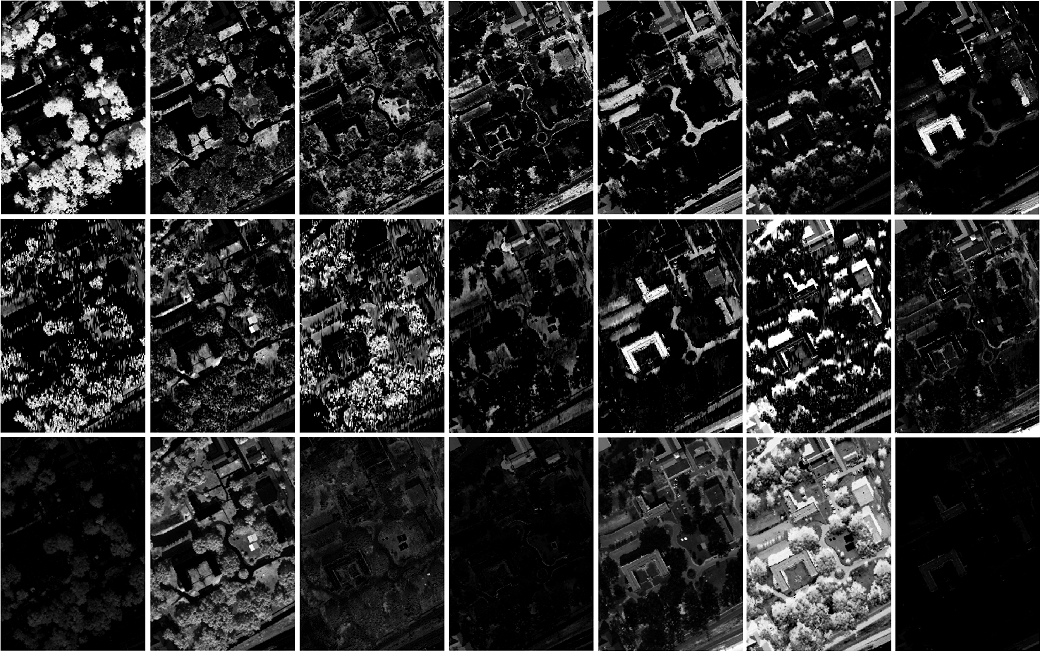}}

\caption{Abundance maps for Mississippi Gulfport dataset. Similarly, the results of EndNet-DMaxD, SCM and SPICE are illustrated in different rows respectively.  Trees  ($\#1$), Grass Pure  ($\#2$), Grass Ground  ($\#3$), Dirt $\&$ Sand  ($\#4$), Road  ($\#5$), Shadow($\#6$) and Building ($\#7$) are identified in columns.}

\label{fig:gulfport}
\end{figure}

Table ~\ref{table:pavia} and ~\ref{table:gulfport} show the SAD performance of these methods. Fig.~\ref{fig:pavia} and~\ref{fig:gulfport} visualize the qualitative fractional abundance performance for three methods: SPICE, SCM and EndNet-DMaxD (VCA, DMaxD and MVSA methods all obtain insufficient abundance results). Remark that $l_{1|2}$-NMF and DgS-NMF scores are not reported since neither their source codes nor previous performance are available on these datasets.

For University of Pavia dataset, the parameters are set to $\beta_1=10$,  $\beta_2=10$ and $\rho=0.01$ for the SCM method. These values are slightly different from the recommended settings in~\cite{zhou2016spatial}, since these configurations yield better visual and quantitative results in our experiments. Similarly, for Mississippi Gulfport, these values are tuned to $\beta_1=10$,  $\beta_2=5$ and $\rho=0.01$. Lastly, the prune threshold value is increased to $10^{-5}$ for SPICE algorithm on Mississippi Gulfport dataset.

\noindent
\textbf{University of Pavia:} Table ~\ref{table:pavia} and Fig.~\ref{fig:pavia} show experimental results on this dataset. As mentioned previously, the scene contains 7 labeled classes including the associations of gravel-bricks and asphalt-bitumen materials. However, we empirically observed that the scene comprises another distinct material, 'Grass', which has a completely different spectral response from 'Meadow' and 'Tree'. These three materials are plotted in Fig.~\ref{fig:endplus}. 

\begin{table*}[t]
\label{tab:load}
\begin{center}

\caption{SAD results on Mississippi Gulfport dataset. Mean and standard deviation are reported. Best results are illustrated in bold. }
\label{table:gulfport}

\begin{tabular}{P{0.8cm} || P{1.4cm} P{1.4cm} P{1.4cm} P{1.4cm}  P{1.4cm}  P{1.4cm} }
  \hline
   \multirow{2}{*}{Endm.} &  \multicolumn{6}{c}{Spectral Angle Distance (SAD) ($\times 10^{-2}$)} \\ \cline{2-7}
      &  VCA  & DMaxD & SPICE & SCM & EndNet-VCA  & EndNet-DMaxD \\ \hline
     \#1   &  17.04 $\pm$$>$9 &  12.71 $\pm$0.0   &  $>$99 $\pm$$>$9 & 15.31 $\pm$0.0 &  3.47 $\pm$1.0 &  \textbf{2.75 $\pm$0.2}    \\
     \#2   & 45.73 $\pm$$>$9 &  12.21 $\pm$0.0   & $>$99 $\pm$$>$9   & 2.07 $\pm$0.0 &  6.32 $\pm$5.2 &  \textbf{1.28 $\pm$0.1}    \\
     \#3   &  25.90 $\pm$1.8 &  26.12 $\pm$0.0   &  $>$99 $\pm$$>$9 & \textbf{1.93 $\pm$0.0} &  9.99 $\pm$8.4 &  3.97 $\pm$0.4    \\
     \#4   &  42.09 $\pm$$>$9 &  34.12 $\pm$0.0   &  $>$99 $\pm$$>$9 & \textbf{4.78 $\pm$0.0} &  13.55 $\pm$$>$9 &  8.31 $\pm$0.7    \\ 
     \#5   &  60.43 $\pm$9.7 &  24.13 $\pm$0.0   &  88.03 $\pm$$>$9 & 29.92 $\pm$0.0 &  23.48 $\pm$$>$9 &  \textbf{2.72 $\pm$0.7}    \\
     \#6   &  39.10 $\pm$9.0 &  47.06 $\pm$0.0   &  $>$99 $\pm$$>$9 & 21.03 $\pm$0.0 &  13.69 $\pm$$>$9 &  \textbf{11.78 $\pm$1.0}    \\ 
     \#7   &  37.42 $\pm$4.7 &  9.28 $\pm$0.0  &  46.60 $\pm$$>$9    & \textbf{6.40 $\pm$0.0} &  8.66 $\pm$2.5 &  11.56 $\pm$0.4    \\ \hline
     Avg.  &  38.25 $\pm$$>$9 &  23.67 $\pm$0.0  &  $>$99 $\pm$$>$9 & 11.65 $\pm$0.0 &  11.31 $\pm$7.9 &  \textbf{6.06 $\pm$0.5}     \\
  \hline \hline

\end{tabular}

\end{center}
\end{table*}

For this reason, we set the optimum endmember number in the scene to 8 just for the proposed method and preserved the default value (i.e. 7) for the baseline methods (these methods yield worse results when we increase the endmember number).

We redefine  $\lambda_2$ and $p$ as $0.01$ and $0.8$ respectively as for the Cuprite dataset, since the data contains a number of correlated materials. We decreased the mask noise to $10\%$ due to the noise level of data and significantly spectral correlations. From the SAD results, it is clear that the proposed method achieves the best performance compared to the baselines and a $3\%$ improvement is introduced over the second best result. 

Of course, the fractional abundance estimation is especially important for this dataset to distinguish the classes. Fig.~\ref{fig:pavia} visualizes the fractional abundance estimation for three methods. It can be seen that our proposed method, EndNet-DMaxD, outperforms the other baseline methods and obtains meaningful abundance results for the materials. This is a gratifying result since it confirms that 'Grass' material in hyperspectral unmixing provides critical improvements to the scene where 'Trees' is repeatedly miscategorized with 'Grass'/'Meadow' by the baseline methods.

\noindent
\textbf{Mississippi Gulfport:} Quantitative and qualitative results for this dataset are reported in Table ~\ref{table:gulfport} and Fig.~\ref{fig:gulfport}. As before, $\lambda_2$ and $p$ values are tuned as $0.1$ and $0.8$, and the mask noise is set to $10\%$. Several materials (i.e. sidewalk, yellow curb, cloth panels and water) were not considered in the computation, since their samples are too scarce in the scene. 

In Table~\ref{table:gulfport}, our proposed method yields the highest overall SAD score over ground truth materials and the quality of abundance estimates in Fig.~\ref{fig:gulfport} is significantly better compared to the baselines. In particular, for 'Tree', 'Building' and 'Shadow', misprediction rates are quite high for the baseline methods, whereas our proposed method obtains nearly optimum results for these materials in the scene.

\section{Conclusion}

In this paper, we proposed a novel method, EndNet, for endmember extraction and hyperspectral unmixing. To this end, we improved and restructured the conventional autoencoder neural network by introducing additional layers and a novel loss function. These modifications enable us to address the common challenges of hyperspectral unmixing such as nonlinearity, sparsity, some of the physical constraints etc. Also, backpropation with a stochastic-gradient based solver is used to optimize the problem and this scales the method for large-scale data, in contrast to more complex derivations/inferences in the literature.

We adapted the SPU algorithm to further improve the estimation of fractional abundances from a scene with the estimated endmembers, even when the estimates of hidden abstracts yield compatible results. This adaptation is achieved by replacing the standard $l2$ norm kernel with a SAD kernel that we have shown is more efficacious.

From the experimental results, we have observed that our proposed method makes significant performance improvements to hyperspectral unmixing domain and achieves outstanding results on well-known hyperspectral datasets. In addition, we selected the DgS-NMF method as a baseline technique in the experiments and we conclusively demonstrated that our method outperforms these type of methods even though they use the spatial priors of a scene in their extraction.

Finally, we should emphasize that the proposed method is the first successful end-to-end learning algorithm based on a neural network that attains superior performance in an unsupervised manner for hyperspectral unmixing problem. That's why, we strongly believe that the findings of our study will provide a basis for further studies into neural network techniques in this domain.

\section{Acknowledgment}
We would like to thank Engin Tola and Ufuk Sakarya for their feedback on the text.

\ifCLASSOPTIONcaptionsoff
  \newpage
\fi



%\bibliographystyle{IEEEtran}
%\bibliographystyle{unsrt}

\bibliographystyle{IEEEtran}
\bibliography{template}
%\bibliography{IEEEabrv,biblio}

\begin{IEEEbiographynophoto}{Savas Ozkan}  is  a senior researcher at TUBITAK Space Technologies Research Institute and he is currently working toward the PhD degree from the  Department  of  Electrical and Electronics Engineering, Middle East Technical University. His research interests include deep learning, image/video retrieval, hyperspectral image processing, generative adversarial networks and biomedical image processing.
\end{IEEEbiographynophoto}

\begin{IEEEbiographynophoto}{Berk Kaya} is a MSc student at ETH Zurich. His research interests are hyperspectral image processing and signal processing.
\end{IEEEbiographynophoto}

\begin{IEEEbiographynophoto}{Gozde Bozdagi Akar}  is  currently a  Professor  with  the  Department  of  Electrical and Electronics Engineering, Middle East Technical University. Her research interests are in face recognition, 2-D and 3-D video compression, multimedia streaming and hyperspectral image processing.
\end{IEEEbiographynophoto}

\vfill



%\vfill


\end{document}